\documentclass[journal]{IEEEtran}
\usepackage{cite}

\ifCLASSINFOpdf
   \usepackage[pdftex]{graphicx}
\else
\fi
\usepackage{amsmath}
\ifCLASSOPTIONcompsoc
  \usepackage[caption=false,font=normalsize,labelfont=sf,textfont=sf]{subfig}
\else
  \usepackage[caption=false,font=footnotesize]{subfig}
\fi

\usepackage{amssymb}
\usepackage{enumitem}
\usepackage{multirow}

\usepackage{xcolor}

\newcommand{\measMat}{\ensuremath{\Phi}}
\newcommand{\basis}{\ensuremath{\Psi}}
\newcommand{\mR}{\mathbb{R}}

\newcommand{\vect}[1]{\ensuremath{\textbf{#1}}}
\newcommand{\sparse}{\ensuremath{\textbf{s}}}
\newcommand{\meas}{\ensuremath{\textbf{y}}}
\newcommand{\tsig}{\ensuremath{\textbf{x}}}
\newcommand{\nsig}{\ensuremath{\tilde{\textbf{x}}}}
\newcommand{\rsig}{\ensuremath{\hat{\textbf{x}}}}

\begin{document}
%
\title{Compressed Sensing for Scalable Robotic Tactile Skins}
%
%
%

\author{Brayden~Hollis,~\IEEEmembership{Student~Member,~IEEE,}
        Stacy~Patterson,~\IEEEmembership{Member,~IEEE,}
        and~Jeff~Trinkle,~\IEEEmembership{Fellow,~IEEE}
\thanks{B. Hollis, S. Patterson, and J. Trinkle are with the Department of Computer Science, Rensselaer Polytechnic Institute, 110 8th Street, Troy, NY, USA,\tt{ hollib@rpi.edu,\{sep,trink\}@cs.rpi.edu}.}
}

\maketitle

\begin{abstract}
The potential of large tactile arrays to improve robot perception for safe operation in human-dominated environments and of high-resolution tactile arrays to enable human-level dexterous manipulation is well accepted. However, the increase in the number of tactile sensing elements introduces challenges including wiring complexity, data acquisition, and data processing. To help address these challenges, we develop a tactile sensing technique based on compressed sensing.  Compressed sensing simultaneously performs data sampling and compression with recovery guarantees and has been successfully applied in computer vision.  We use compressed sensing techniques for tactile data acquisition to reduce hardware complexity and data transmission, while allowing fast, accurate reconstruction of the full-resolution signal. For our simulated test array of 4096 taxels, we achieve reconstruction quality equivalent to measuring all taxel signals independently (the full signal) from just 1024 measurements (the compressed signal) at a rate over 100Hz. We then apply tactile compressed sensing to the problem of object classification. Specifically, we perform object classification on the compressed tactile data based on a method called compressed learning.
We obtain up to 98\% classification accuracy, even with a compression ratio of 64:1.
\end{abstract}


%
\IEEEpeerreviewmaketitle

\section{Introduction}
%
%
%
%
\IEEEPARstart{F}{or} robots to reliably and safely function in unstructured environments, they need to perceive and react to the world around them.  
Sensors that operate without contact, such as LIDAR and sonar, are useful in building geometric models and tracking moving objects, but robots need information about contact to do physical tasks. Tactile sensors range from simple sensors that measure only the locations of contacts to more sophisticated tactile sensors that measure surface properties, such as temperature, force, roughness, conductivity, and mechanical stiffness.   Such tactile data, when assimilated with a gross geometric model of the environment, can greatly increase a robot's understanding.

An important area of research in tactile sensing is tactile systems that cover large portions of robot bodies.  These systems, also called \emph{tactile skins}, provide improvements in a robot's awareness and manipulation abilities.  For example, Ohmura \emph{et al.}~\cite{box} demonstrate a robot lifting a heavy box with the use of a tactile skin, a task beyond the robot's capability with traditional methods. Another example is Bhattacharjee \emph{et al.}~\cite{Bhattacharjee2012} in which a tactile skin on a robot arm is used to classify objects as movable or fixed, allowing the robot to reach through clutter. A number of other skills, for instance, safe human-robot interaction, 
can also benefit from tactile sensors on the surfaces of various links.

There are numerous challenges to fully realizing tactile skins.  To cover the surfaces of a robot at useful resolutions requires a large number of individual tactile sensing elements, also called \emph{taxels}, on the order of thousands to millions~\cite{dahiya10tactile}.  
The massive amounts of data generated by these taxels needs to be gathered and processed at high rates, up to $1kHz$ for fine force control~\cite{dahiya10tactile}. The difficulty of quickly managing this large amount of data  is increased by the limited room for the wiring of these systems, especially in skins designed separately from the robot as add-ons. 
Most current systems rely on fast hardware for gathering the data directly from every taxel~\cite{ohmura06sensor, schmitz11icub, mitt11hex}, which limits the total number of taxels that can be embedded in the skin while supporting high data rates.  Other systems perform local feature extraction from the tactile data, such as the center of contact, prior to transferring it to the main processing center~\cite{riman,schmitz11icub}.  This allows for more taxels on the skin, but it limits the information available to the robot for completing its tasks.

 

In this work, we present solutions for real-time sensing in large-scale tactile skins that support acquisition and processing of the full tactile data signal.
We first consider the problem of efficiently collecting tactile data and transferring it to a processing center (i.e., the robot's brain) at high rates, a problem that we call the \emph{data acquisition problem}.
Our proposed solution is based on the theory of compressed sensing~\cite{cs}. Compressed sensing is a popular technique for signal processing in which a signal is compressed in hardware during the sampling process.
Provided the original signal is sparse in some basis, it can be efficiently reconstructed from its compressed version with little to no signal loss. 

We first demonstrate that tactile sensor data is amenable to compressed sensing by identifying bases under which this data is sparse.  We then identify sampling and compression operators that are suited for  tactile system hardware.  
Finally, we implement different methods for signal reconstruction from the compressed data that provide accurate reconstructions at high rates. 
Our approach opens the door to the construction of large-scale tactile skins that supports accurate full data acquisition at high speed, with reduced wiring complexity. 
Our evaluations with simulated data show that our compressed sensing-based approach, with a compression ratio of 4~to~1, can achieve higher signal acquisition accuracy than full data acquisition of noisy sensor data of 4096 taxels, with a reconstruction time under $5ms$ per time step.



While our approach addresses the scalability challenges of tactile data acquisition, the sheer size of the data that is generated may still be prohibitive for certain applications.  Thus, we also explore the possibility of using the compressed data directly. Specifically, we consider the application of \emph{tactile object classification}.
Tactile object classification utilizes tactile signals to distinguish between different objects or object classes.  A motivating scenario is a robot searching for specific items on a shelf that is out of view of its cameras.  Using tactile object classification, the robot can feel around until it finds the correct object rather than  removing  objects from the shelf for visual inspection. Not only is removing objects for visual inspection less efficient with regards to time, grasping out-of-sight objects is challenging, especially if there is no tactile feedback.


We develop a method for tactile object classification that uses compressed tactile signals.
Our approach is based on theoretical results by Calderbank \emph{et al.}~\cite{Calderbank2009} called \emph{compressed learning}, which define mathematical conditions under which classification performed using compressed signals achieves nearly the same accuracy as classification performed using the full signals. 
Our tactile signals are compressed from snapshots of data obtained from a simulated array of taxels pressed onto objects from above, similar to what might be obtained in the cupboard example, and we use a soft-margin support vector machine for the object classification.
With this approach, the data dimensionality can be reduced in hardware during data acquisition, rather than as a pre-processing step in classification.  This, in turn, reduces computational needs for both training and classification with minimal impact on the accuracy of the classifier. 
Further, in applications for which the full signal is required, it can be recovered from the compressed signals.
In evaluations, our method is able to classify among 16 objects with approximately 95\% accuracy using compressed signals with a compression ratio of 64 to 1. 
 


The rest of this paper is organized as follows. In Section~\ref{rwts}, we present related work on tactile data acquisition and tactile object classification.  We then present background on compressed sensing and classification in Section~\ref{background.sec}.  In Section~\ref{prob.sec}, we explain the problems we address in tactile sensing and object classification.  Our data sets used for developing and evaluating our methods are discussed in Section~\ref{data.sec}.  We detail our solutions for tactile data acquisition and tactile object classification in Section~\ref{sol.sec}, and we present empirical evaluations of our solutions in Section~\ref{eval.sec}. Finally, we conclude in Section~\ref{concl} with final remarks and topics for future work.

We note that a preliminary version of our work on compressed sensing for tactile data acquisition appeared in~\cite{Hollis2016}.  
This paper expands on our previous work through exploring additional methods that enable greater compression of the tactile signal.  Further, we propose an additional compression operator, with a hardware implementation that may be more suitable for some applications. Finally, this paper extends beyond using compressed sensing purely for tactile data acquisition by applying compressed learning to the task of tactile object classification.

\section{Related Work} \label{rwts}
We now present the most relevant related work on tactile data acquisition and tactile object classification.

\subsection{Tactile Data Acquisition}
There have been a number of tactile skins developed in the past few decades and here we present a representative subset.  

The majority of systems perform full data acquisition and rely on having a limited number of taxels connected via high capacity networks to achieve fast data acquisition~\cite{ohmura06sensor, schmitz11icub, mitt11hex}.  
These systems typically have local micro-controllers that each serially sample a subset of the taxels. 
The micro-controllers share a high capacity network to transfer the data to the processing center. This method has been implemented on a skin with 4200  taxels with a sampling rate of up to $100Hz$~\cite{schmitz11icub}.  However, these systems are constrained by the data capacity of the networks, so
to scale the number of taxels while maintaining high acquisition rates will require improved hardware capabilities or additional high capacity networks.
Adding additional networks is undesirable due to limited spacing for wires~\cite{dahiya10tactile}.  In contrast, our approach reduces the amount of data transferred across the networks, allowing larger numbers of taxels for a given network.

An alternate approach to full data acquisition is to use local processing of sensor data within small clusters of sensors~\cite{Mukai2008,schmitz11icub}.
This approach increases  tactile data processing rates by only transferring aggregate information to the processing center. 
Local processing like this suffers from a loss in data quality or a reduction in the versatility of the system because data features must be determined \emph{a priori}.
In contrast, our solution needs fewer measurements and can recover the full signal when needed.

One of the most recent approaches is event-triggered acquisition~\cite{Bergner2015b}.
In such methods, each taxel transmits its sensor reading only when this reading changes in some detectable way, usually determined by a threshold. 
The processing center stores a record of each taxel's transmissions, and by doing so, can maintain an approximation of the full tactile data set.
This approach can reduce the amount of data that is transferred to the processing center; however, it requires some on-board processing for each taxel.
Further, the event detection thresholds must be set \emph{a priori}, which can limit the adaptability of the sensing system.
Finally, when a large number of taxel values change simultaneously, the system may transfer more data than traditional full data acquisition due to the need of transferring an identifier along with each taxel reading.
Our approach, on the other hand, requires minimal processing in each taxel, and the data generation rate remains consistent regardless of the types of contact.

\subsection{Tactile Classification}
Tactile object classification is an active area of research~\cite{Russell2000,Heidemann2004,Schopfer2007,Schopfer2009,Takamuku2008,Schneider2009,Gorges2010,Hosoda2010,Pezzementi2011,Schmitz2014}. 
A number of these approaches use the full tactile data gathered from contact with the object~\cite{Jimenez1997,Hosoda2010,Schmitz2014}; however, these approaches are computationally expensive and risk impracticality due to the curse of dimensionality for anything but small-scale tactile skins.
To address these limitations, a number of methods reduce the dimensionality of the data before, or as part of, the machine learning process.  Some manually reduce the dimension by selecting features such as average force and contact area~\cite{Schopfer2009,Duarte2011,Bhattacharjee2012,Drimus2014}. Others use automated methods to reduce the dimension, for instance, Self-Organizing Maps~\cite{Takamuku2008,Gorges2010} and Principle Component Analysis~\cite{Heidemann2004,Schopfer2007,Pezzementi2011}.  
In our approach, the dimension of the data is reduced as part of the acquisition process done in hardware.  Because the dimension reduction is not done computationally, it saves processing time and effort.



\section{Background} \label{background.sec}

In this section, we describe the theory of compressed sensing.  We then give a brief overview of support vector machines (SVMs).  
Lastly, we present background and theoretical results for compressed learning.

\subsection{Theory of Compressed Sensing} \label{cs.sec}

In compressed sensing, a signal, for example, force readings from a tactile array at a given time,
 is simultaneously measured and compressed by taking linear combinations
of the signal components.  
We call the signal to be measured the \emph{true signal} $\tsig \in \mR^n$.
The \emph{compressed signal} $\meas \in\mR^m$, with elements $y_i$, is obtained from
the true signal as follows:
\begin{equation} \label{system.eq}
    \meas = \measMat \tsig,
\end{equation}
where $\measMat = [\measMat_{ij}]$ is the $m \times n$ \emph{measurement matrix}.
If $m < n$, (\ref{system.eq}) is under-determined, and, in general, $\tsig$ cannot be recovered.  

In compressed sensing, one considers a restricted set of signals that are sparse in some representation basis.
Formally, a signal $\tsig$ is $k$-sparse in a \emph{representation basis} $\basis \in \mR^{n\times p}$ if there exists an $\sparse \in \mR^p$ that is $k$-sparse, meaning $\sparse$ has at most $k$ non-zero entries, such that $\tsig = \basis  \sparse$. We refer to $\sparse$ as the \emph{sparse signal}.
The theory of compressed sensing provides conditions under which such sparse signals can be recovered from fewer than $n$  measurements.
One such condition is the \emph{restricted isometry property}, which is defined as follows.
\newtheorem{myDef}{Definition}
\begin{myDef} \label{RIP.def}
A matrix $A$ satisfies the $k$-\emph{restricted isometry property} ($k$-RIP) if there exists a $\delta \in (0,1)$ such that 
\begin{equation} \label{RIP.eq}
    (1-\delta)\|\sparse\|_2^2 \leq \|A\sparse\|_2^2 \leq (1+\delta)\|\sparse\|_2^2,
\end{equation}
holds for all $k$-sparse vectors $\sparse$.
\end{myDef}

If $\tsig$ is $k$-sparse in a representation basis $\basis$, and the matrix ${A=\measMat \basis}$ satisfies the $2k$-RIP, then $\tsig$ can be recovered exactly~\cite{cs}.


In general, determining whether a given matrix $A$ satisfies RIP is NP-Hard~\cite{cs}. However, several classes of matrices have been
shown to satisfy $k$-RIP with high probability, when $m \geq 0.28k\log{\frac{p}{k}}$~\cite{cs}. For example, if $A$ is a random matrix with entries drawn from a Gaussian distribution, then with high probability, $k$-RIP will be satisfied when $m~=~O(k \log (p/k)/\delta^2)$~\cite{baraniuk2008simple}. The \emph{compression ratio} is the reduced fraction $n/m = n_r/m_r$ written as $n_r$:$m_r$.
In practice, other classes of matrices have been applied to compressed sensing with similar compression ratios, even when $k$-RIP cannot be theoretically guaranteed~\cite{Duarte2011}.

The sparse signal can be recovered from the compressed signal by solving an optimization problem of the form,
\begin{equation} \label{cs1.eq}
 \underset{\sparse \in \mR^{^p}}{\text{minimize}}~\| \sparse \|_0 ~~~\text{subject to}~\measMat \basis \sparse = \meas
\end{equation}
Here $\| \cdot \|_0$ denotes the $\ell_0$ pseudonorm, i.e., the number of non-zero components.
It is intractable to solve problem (\ref{cs1.eq}) directly; however, efficient alternate approaches to find the solution have been derived~\cite{cs}.
Once the solution $\hat{\sparse}$ of (\ref{cs1.eq}) is obtained, the \emph{reconstructed signal} $\rsig$ is computed as $\rsig = \basis \hat{\sparse}$.

In this work, we consider a variation on the compressed sensing problem where measurements may not be exact.
In this case, the compressed signal is given by
\begin{equation}
   \meas = \measMat \basis \sparse + \nu, 
\end{equation}
where $\nu \in \mR^{m}$ is the measurement noise.  This noise can model both errors in the individual sensor readings, as well as in the measurement acquisition.  

The sparse signal can be recovered using Basis Pursuit Denoising (BPDN)~\cite{Chen2001}. In BPDN, one solves a convex relaxation of (\ref{cs1.eq}) that also incorporates measurement noise, 
\begin{equation} \label {bpdn.eq}
 \underset{\sparse \in \mR^{^p}}{\text{minimize}}~\textstyle \frac{1}{2} \displaystyle \| \measMat \basis \sparse - \meas \ \|^2_2 + \lambda \| \sparse \|_1,
\end{equation}
where $\lambda$ is a regularization parameter that is used to tune the level of sparsity of the solution. It has been shown that BPDN produces solutions $\hat{\sparse}$, that are optimal, i.e., $\|\sparse-\hat{\sparse}\|_2 = \vect{0}$, or near-optimal in a wide variety of settings~\cite{TW10}.
Further, since problem (\ref{bpdn.eq}) is convex, it can be solved efficiently using one of many algorithms for convex optimization, as well as variants
developed specifically for compressed sensing.

Many signals in real-world applications are only \emph{approximately sparse}, meaning that, while they are not sparse, they can be well approximated by sparse signals.  
The quality of the sparse approximation is typically evaluated as $\|\sparse-\sparse_k\|_2$, where $\sparse_k$ is the best  $k$-sparse approximation of the approximately sparse signal $\sparse$.  
Since  the solution to (\ref{bpdn.eq}) is a sparse signal $\hat{\sparse}_k$,  if $\|\sparse - \sparse_k\|_2$ is small, then the recovered signal will be a good approximation to the original signal.
The best $k$-sparse approximation is the signal that keeps the $k$ entries in $\sparse$ with largest magnitude and sets the rest to zero~\cite{cs}.
Using this idea, one can determine the approximate sparsity of a signal, i.e.,
\begin{equation} \label{sparsity.eq}
    \tilde{k} = |\{s_i \in \sparse | s_i > \tau \}|,
\end{equation}
where $\tau$ is a small threshold.

\subsection{Support Vector Machines} \label{svm.sec}
SVMs are a popular classification tool in machine learning.  They classify \emph{observations}, which are data instances (e.g., tactile sensor signals) representing things being classified (e.g., objects), by finding a hyperplane that separates the \emph{training set}, a set of observations with known classes, into its two classes, such that the distance between the hyperplane and the closest observations is maximized~\cite{Burges1998}.  This distance is called the \emph{margin} and the closest observations are called \emph{support vectors}, as they are the  observations that  determine the classifier.  To find the hyperplane, one solves the following minimization problem,
\begin{equation} \label{svm.eq}
    \begin{aligned}
        & \underset{b,\vect{w}}{\text{minimize}}
        & & \frac{1}{2}\vect{w}^T\vect{w} \\
        & \text{subject to}
        & & \ell_i(\vect{w}^T\vect{x}_i + b) \geq 1\\
        & & &\text{for }i = 1, \dots, N,
    \end{aligned}
\end{equation}
where $N$ is the number of observations; $\vect{x}_i \in \mR^n$, for $i = 1$ to $N$, are the observations; $\ell_i \in \{-1,1\}$, for $i = 1$ to $N$, are the class labels; $\vect{w}$ is a vector orthogonal to the separating hyperplane; and $b=-\vect{w}^T\vect{x}_0$, for any point $\vect{x}_0$ on the hyperplane.  This optimization problem is a convex quadratic program, a well studied class of problems with many implemented solvers.  Once (\ref{svm.eq}) is solved for $\hat{\vect{w}}$ and $\hat{b}$, to classify a point $\vect{x}'$, one simply computes $\text{sign}(\hat{\vect{w}}^T\vect{x}'+\hat{b})$.

Often it is not possible to completely separate the data with a hyperplane. In such cases, SVMs can been modified to allow some observations to be misclassified as follows~\cite{Burges1998}:
\begin{equation}\label{softsvm.eq}
    \begin{aligned}
        & \underset{b,\vect{w}}{\text{minimize}}
        & & \frac{1}{2}\vect{w}^T\vect{w}  + C\sum_{i=1}^N\xi_i\\
        & \text{subject to}
        & & \ell_i(\vect{w}^T\vect{x}_i + b) \geq 1-\xi_i\\
        & & & \xi_i \geq 0\\
        & & &\text{for } i = 1, \dots, N,
    \end{aligned}
\end{equation}
where $\xi_i$ is the amount the $i$th observation violates the margin, known as the \emph{hinge loss}, and $C$ is a parameter to balance between maximizing the margin and reducing the hinge loss.  This modification defines soft-margin SVMs.

When training soft-margin SVMs, cross-validation is used to tune the parameter $C$.  Cross-validation separates the training set of (\emph{observation}, \emph{label}) pairs into a \emph{development set} and a \emph{validation set}. The development set is used to train multiple SVMs.  All the trained models are then evaluated with the validation set.  The model that performs the best on the validation set is then retrained using the full training set and is the final learned classifier.

While maximizing classification accuracy (the percent of classes correctly labeled) is the true objective of SVMs, SVMs can also be evaluated by the \emph{expected hinge loss} $H_D(\vect{w}^+)$ over the probability distribution $D$ of all possible observations, where $\vect{w}^+\in\mR^{n+1}$ is the vector generated by concatenating $\vect{w}^*$ and $b^*$.  More formally,
\begin{equation}
    H_D(\vect{w}^+) = \mathbb{E}_D\bigg[\max\{0,1-y({\vect{w}^*}^T\vect{x} + b^*)\}\bigg].
\end{equation}
$\mathbb{E}_D[\cdot]$ is the expectation over $D$.  In general, one desires the expected hinge loss to be as small as possible.

\subsection{Compressed Learning Theory} \label{clt.sec}
Calderbank \emph{et al.} proposed classification using compressed signals, a technique that they have called compressed learning~\cite{Calderbank2009}.
Specifically, they show that, with high probability, a soft-margin SVM trained on compressed signals, generated from a measurement matrix that satisfies RIP, has an expected accuracy similar to that of the best linear classifier of the full signals.  This is formalized in the following theorem.
\newtheorem{thm}{Theorem}
\begin{thm}[Thm. 3.1~\cite{Calderbank2009}] \label{cl.thm}
Let $D$ be a distribution of $k$-sparse vectors $\tsig_i \in \mR^n$ such that  for all $i$, ${\|\tsig_i\|_2 \leq R}$, where $R$ is a known upper bound. Further, assume that for each $\tsig_i$ there is a label $\ell_i \in \{-1,1\}$.  Let $\measMat\in\mR^{m~\times~n}$ be a measurement matrix that satisfies $2k$-RIP with constant $\delta$.  Additionally, let 
\begin{equation*}
    S_\measMat = \{(\measMat \tsig_1,\ell_1),...,(\measMat \tsig_N,\ell_N)\}
\end{equation*}
be a set of i.i.d. labeled instances compressively sampled from $D$, and let $\vect{z}_{S_\measMat}\in\mR^m$ be the linear classifier from the soft-margin SVM trained on $S_{\measMat}$.  Finally, let $\vect{w}_0\in\mR^n$ be the best linear classifier with low expected hinge loss over $D$, $H_D(\vect{w}_0)$, and large margin (hence small $\|\vect{w}_0\|_2$).  Then with probability $1-2\rho$ over $S_\measMat$:
\begin{equation} \label{bound.eq}
    H_D(\vect{z}_{S_\measMat}) \leq H_D(\vect{w}_0) + O\Bigg(\|\vect{w}_0\|_2 \bigg(R^2 \delta + \frac{\log(\frac{1}{\rho})}{N}\bigg)^{\frac{1}{2}}\Bigg).
\end{equation}
\end{thm}

This theorem means that, in general, the expected hinge loss, when training and classifying using the compressed signals, is within a bound of the best possible classifier using the full signals.  
The bound is essentially dependent on the compression ratio and the size of the training set, since $R$, $\vect{w}_0$, and $\rho$ are fixed by the problem. The variable $\delta$ relates to the data compression, with $\delta$ typically increasing as the amount of compression increases, i.e., the compression ratio decreases.  Thus, (\ref{bound.eq}) implies greater compression leads to less confidence in the classifier's accuracy, which is to be expected.  
Also, as expected, the accuracy depends on the training set size $N$. As $N$ increases, the accuracy of the compressed classifier approaches that of the optimal linear classifier.

Theorem~\ref{cl.thm} addresses classification with compressed signals obtained when the true signals $\tsig$ are themselves sparse.
If the true signals are not sparse, but are sparse in some representation basis $\basis$, then Theorem 1 still applies, provided 
$\measMat \basis$ satisfies $2k$-RIP with constant $\delta$.
An important point to note is that, unlike in compressed sensing, for compressed learning it is not necessary for this representation basis to be known.  

\section{Problem Statement} \label{prob.sec}
 We model the tactile skin as an array of $n$ individual taxels arranged in a $\sqrt{n} \times \sqrt{n}$ grid. 
The tactile data (the true signal) is a vector-valued signal, $\tsig(t) \in \mR^n$, where each element $\tsig_i(t)$ corresponds to the force applied to $i^{th}$ taxel at time $t$. 
The sensors do not measure the applied force exactly, but rather, they measure the \emph{sensor signal}, $\nsig(t) = \tsig(t) + \mathbf{\epsilon}(t)$, where $\mathbf{\epsilon}(t) \in \mR^n$ is the sensor noise.

\subsection{Requirements for Data Acquisition in Tactile Skins} \label{acq_req.sec}

The tactile data acquisition problem is the problem of transmitting the full tactile data $\tsig(t)$ from the tactile skin to the processing center.  
The signal is continually updated as contact with the robot changes, and thus, 
it is desirable to acquire snapshots of  $\tsig(t)$ at a high rate, ideally approaching $1kHz$.  As discussed in Section~\ref{rwts}, approaches that sample each taxel directly have scalability limitations.
We propose to address these limitations using  techniques from compressed sensing.  In our approach, in each time step, the full sensor signal $\nsig(t)$ is simultaneously measured and compressed
into $m$ measurements, forming the compressed signal $\meas(t) = \measMat\nsig(t)$.  
This compressed signal is transmitted to the processing center, where a close approximation $\rsig(t)$ to the true signal $\tsig(t)$ is reconstructed from $\meas(t)$.

To apply compressed sensing to tactile data acquisition, we need to address the following three components:
\begin{itemize}[leftmargin=0cm,itemindent=.3cm,labelwidth=\itemindent,labelsep=0cm,align=left]
\item{\textbf{Identification of a representation basis $\basis$.}} We must identify a basis under which the true signals  from our data sets are sufficiently sparse.
A greater level of sparsity means that the signals can be accurately reconstructed from fewer measurements, as described in Section~\ref{cs.sec}.  This will  speed up the acquisition process because less data needs to be gathered and transferred to the processing center.  
The representation basis should also be universal, 
meaning that the signals from many kinds of contact and manipulation, for example, the robot being bumped or the robot carrying an object,
should all have sparse representations in the basis.  
This universality will allow the same hardware and algorithms to be used for tactile data acquisition in a wide variety of  applications.

\item \textbf{Identification of a measurement matrix $\measMat$ that is compatible with the representation basis and hardware.}  As mentioned in Section~\ref{cs.sec}, 
for the true signal to be recoverable from the compressed signal, the matrix $\measMat \basis$ must possess certain mathematical properties, for instance, RIP.  
For tactile sensing, we have a further restriction that the measurement matrix be hardware compatible, meaning that the linear combinations over the sensor signal can be performed in hardware across the tactile array. 
This means only $m < n$ measurements are required from the hardware.
Having the matrix be hardware compatible is important for the scalability of the system since hardware compression has  the potential to reduce the amount of wires and improve the wiring layout within the skin.
It also requires less bandwidth between the skin and the processing center.

\item \textbf{Selection and implementation of a fast signal reconstruction algorithm.} 
The reconstruction algorithm needs to be able to reconstruct the signal $\rsig(t)$ from
the compressed signal $\meas(t)$ in real time so the signal can be used in control loops.
To do this, we use BPDN, as described in (\ref{bpdn.eq}), for which there are a variety of efficient algorithms~\cite{cs}.
We must select an algorithm that can be implemented in the robot's hardware, which we assume is a centralized processing system that may include a GPU.  
Many BPDN algorithms are iterative; in each iteration, the reconstructed solution is refined, approaching the optimal solution to (\ref{bpdn.eq}) in the limit of infinite iterations.
Thus, for these algorithms,  there is a trade-off between the time to reconstruct the signal and the accuracy of the reconstructed signal.  We seek algorithms
that can produce accurate reconstructions in just a few iterations so that processing time is as small as possible.
\end{itemize}

In Section~\ref{acq_solution.sec}, we describe our approach to each of these three components.

\subsection{Requirements for Tactile Object Classification} \label{class_req.sec}
In tactile object classification, a tactile signal is used to classify an object, where each class is a known object or set of objects.  
To perform tactile object classification in a system that uses our compressed sensing approach, one 
can use an existing classification approach on the recovered signals.  However, as the number of taxels increases, so too does the computational complexity
of the classification procedure.  We propose to reduce this complexity by performing object classification using the compressed signals directly, leveraging the
theory of compressed learning that was presented in Section~\ref{clt.sec}.

 To generate each observation, we collect a snapshot of the compressed signal $\meas(t)$ at a single time instance $t$ during contact with the object.  The full details 
 of the observation generation are given in Section~\ref{to_exp.sec}. Let $\nsig_i$ be the sensor signal for observation $i$, and let $\vect{y}_i = \measMat \nsig_i$ denote the
 corresponding compressed signal. 
For each observation $\vect{y}_i$, the objective is to determine the appropriate label $\ell_i$ that corresponds to the object that generated the observation.

To solve this problem, we must develop a classifier that can accurately label objects based on their compressed signals.  
We base our approach on the theory of compressed learning described in Section~\ref{clt.sec}, which shows that classification on compressed signals using soft-margin SVMs (described in Section~\ref{svm.sec}),
can perform nearly as well as soft-margin SVMs using the true signals.  The original result applies only to binary classification, whereas we consider classification over more than two objects.  Thus, our solution must utilize a multi-class classifier while still satisfying the theoretical assumptions of compressed learning.

We present our compressed learning-based approach  for tactile object classification in Section~\ref{class_solution.sec}.

 



\section{Data Set Description} \label{data.sec}

We generate tactile data using our BubbleTouch simulator~\cite{bubbletouch}.  BubbleTouch represents taxels as rigid spheres suspended in space by spring and damper pairs, one pair per sphere.
Contact interactions between objects and taxels are assumed to be quasistatic to avoid simulation instabilities that commonly arise in the simulation of dynamic systems with intermittent contact.  Tactile arrays of any shape and distribution can be simulated  by simply creating a rigid substrate body with that shape and attaching the bases of the springs to it in the desired pattern. For this work, we created seven planar square grid arrays with dimensions $256mm$ by $256mm$ of 4096, 1024, 256, 64, 16, 4, and 1 taxel(s).

\begin{figure}
\centering
\includegraphics[width=.77\linewidth]{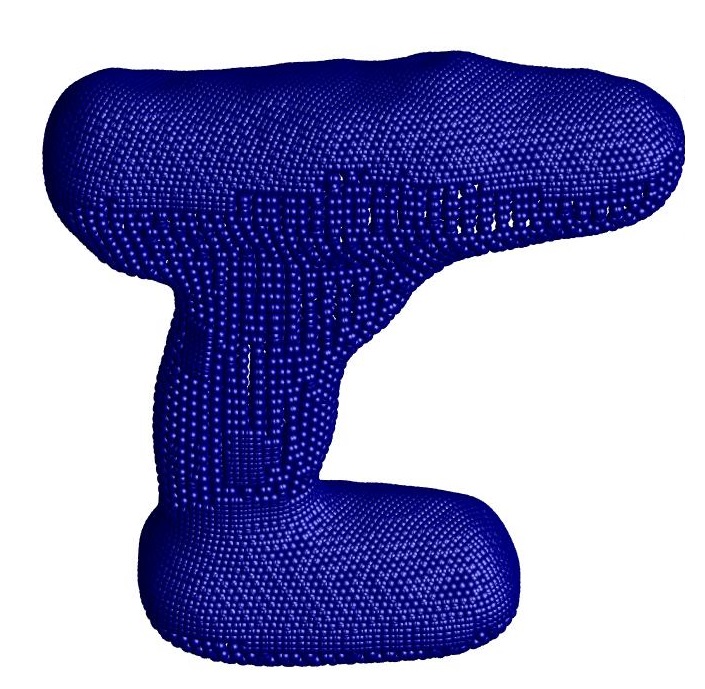}
\caption{A union-of-spheres model generated from the Yale-CMU-Berkeley (YCB) drill.}
\label{drillmodel.fig}
\end{figure}

Our objects for generating our tactile data consist of the banana, cup, drill, large clamp, and mustard bottle, obtained from the  Yale-CMU-Berkeley (YCB) Object Set~\cite{Calli2015}.  We also use a golf ball, racquetball, volleyball, basketball, cracker box, cereal box, jello box, granola box, gravy can, tuna can, and salmon can, generated as primitive geometric shapes of the appropriate dimensions.  To simplify collision detection between the tactile array and objects, each object is approximated by a union of spheres. To convert a YCB object to a union-of-spheres model, we used the vertices from the UC Berkley Poisson mesh models as the sphere centers.  For each object of $V$ vertices, all of the spheres were assigned a radius $r$ equal to twice the mean distance between all vertices $v_i, i=1,..,V$, and their nearest neighboring vertex, i.e., $r = \frac{2}{V}\sum_{i=1}^V \min_{j \neq i} \|v_i-v_j\|_2$.
An example of a union-of-spheres model for the YCB drill is shown in Fig. \ref{drillmodel.fig}.  For the ellipsoids, boxes, and cylinders representing the other objects, we manually designed the union-of-spheres models.

\begin{figure*} 
\centering
\begin{tabular}{cccc}
\subfloat[banana]                {\includegraphics[width = 1.4in]{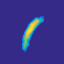}\label{banana.fig}} &
\subfloat[cup]                   {\includegraphics[width = 1.4in]{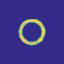}\label{cup.fig}} &
\subfloat[drill]                 {\includegraphics[width = 1.4in]{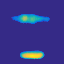}\label{drill.fig}} &
\subfloat[clamp]         {\includegraphics[width = 1.4in]{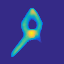}\label{clamp}}\\
\subfloat[mustard bottle - side] {\includegraphics[width = 1.4in]{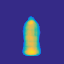}\label{mustard_side.fig}} &
\subfloat[mustard bottle - up]   {\includegraphics[width = 1.4in]{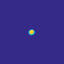}\label{mustard_up.fig}} &
\subfloat[cracker box]     {\includegraphics[width = 1.4in]{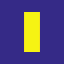}\label{cracker.fig}} &
\subfloat[cereal box]      {\includegraphics[width = 1.4in]{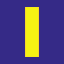}\label{cereal.fig}}\\
\subfloat[jello box]       {\includegraphics[width = 1.4in]{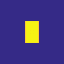}\label{jello.fig}} &
\subfloat[granola box]{\includegraphics[width = 1.4in]{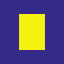}\label{granola.fig}} &
\subfloat[racquetball]     {\includegraphics[width = 1.4in]{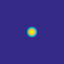}\label{racquetball.fig}} &
\subfloat[volleyball]      {\includegraphics[width = 1.4in]{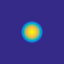}\label{volleyball.fig}}\\
\subfloat[basketball]      {\includegraphics[width = 1.4in]{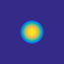}\label{basketball.fig}} &
\subfloat[gravy can]       {\includegraphics[width = 1.4in]{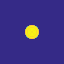}\label{gravy.fig}} &
\subfloat[tuna can]        {\includegraphics[width = 1.4in]{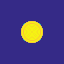}\label{tuna.fig}} &
\subfloat[salmon can]      {\includegraphics[width = 1.4in]{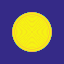}\label{salmon.fig}}
\end{tabular}
\caption{Example noise-free tactile images for each object.  The color scale goes from dark blue (no contact force) to yellow (contact force of $0.01N$).}
\label{objectlist.fig}
\end{figure*}

To generate tactile data, each object was placed in a stable configuration on a rigid horizontal support plane and touched from above by the tactile array.  Each touch was performed by initially positioning the substrate of the array far enough above the object to avoid contact.  From there, the substrate was translated downward, causing the taxels to contact the object and move relative to the substrate (deforming the springs and dampers).  After a pre-specified downward motion of the substrate, the substrate was translated parallel to the support plane for a pre-specified amount of time.  During these simulations, the combined spring and damper forces at all the taxels were taken as the true signal $\tsig(t)$, at a frequency of $1kHz$.  Example true signal tactile images from 16 objects (we treat the two different orientations of the mustard bottle as separate objects) are shown in Fig.~\ref{objectlist.fig}. To approximate the noise in real tactile sensors, random zero-mean Gaussian noise with standard deviation of $0.001N$ (equal to $5\%$ of the signal range) was added to each taxel reading.  Values outside a taxel's range ([$0, \ 0.02$]$N$) were clipped to the boundary.  This created the sensor signal $\nsig(t)$.  

\subsection{Tactile Data Acquisition Data Set} \label{cs_exp.sec}

For our tactile data acquisition work, we consider five objects that generate different types of contact: the golf ball for a small number of taxels in contact, the granola box (Fig. \ref{granola.fig}) for a large number of taxels in contact, the drill (Fig. \ref{drill.fig}) for multiple contact points, the cup (Fig. \ref{cup.fig}) for a topological contact different from the other objects, and the clamp (Fig. \ref{clamp}) for a non-convex contact.  We used the full trajectory data of the 4096-taxel array, which consists of 4200 time steps.  Note that the parallel translation of the object activates taxels throughout the array.
This allows us to verify whether contact location is a factor in the accuracy of our data acquisition approach.

\subsection{Tactile Object Classification Data Set} \label{to_exp.sec}

We use all of the objects whose tactile images are shown in Fig.~\ref{objectlist.fig} for our tactile object classification data sets. The sensor signal for an observation is recorded at the last time step prior to horizontal object translation.  
We collected seven data sets of sensor signals, one for each array of different numbers of taxels (4096, 1024, 256, 64, 16, 4, and 1).  
We generate our compressed signal data sets from the 4096 taxels array data set.  The remaining data sets are used to compare the accuracy of classification using compressed signals
versus classification using tactile arrays with fewer, larger taxels. We compare our approach to lower resolution tactile arrays because successful classification with fewer taxels would also reduce hardware complexity and address the curse of dimensionality.

In each data set, for each of the 16 objects, we generated 360 observations through systematic offsets from the nominal starting configuration of the array.  The offsets were $(0,2,4,6,8,10)mm$ along the rows of the array, $(0,2,4,6,8,10)mm$ along the columns of the array, and $(0^\circ,5^\circ,10^\circ,15^\circ,20^\circ,25^\circ,30^\circ,35^\circ,40^\circ,45^\circ)$ rotations about an axis normal to the array. This yielded a total of 5,760 observations per data set.

\section{Solution} \label{sol.sec}
We first describe our tactile data acquisition system that solves the data acquisition and object recognition problems stated in Section~\ref{acq_req.sec}.  We then explain our extension for tactile object classification.

\subsection{Tactile Data Acquisition Solution} \label{acq_solution.sec}

Recall, our goal is to take compressed measurements $\meas$ of a noisy tactile signal, i.e., $\meas = \measMat \nsig$, and then, from the compressed measurements, reconstruct a signal $\rsig$ that approximates the true tactile signal $\tsig$.  To do this we must find 1) a representation basis $\basis$ in which the true signal is sparse, 2) a measurement matrix $\measMat$ that can be implemented in hardware and such that $A = \measMat \basis$ satisfies $k$-RIP, and 3) a algorithm that quickly and accurately reconstructs the tactile signal.

\subsubsection{Representation basis $\basis$} \label{basis.sec}
To determine a suitable representation basis, we investigate a set of bases to determine which yields the sparsest sparse signal for our data sets.  We select this set from bases used in image processing because the array pattern in our tactile sensor corresponds well with image pixel arrays.  This similarity between images and tactile arrays have been exploited for other tactile processing tasks~\cite{dahiya13tactilereview}.

We investigate two classes of bases: the discrete cosine transform (DCT) and the Daubechies wavelet transforms. DCTs are used in the original JPEG compression scheme, while wavelets are used in JPEG2000.  
We also  consider the contourlet transform~\cite{Do2001a}.  Contourlets are specifically designed for two-dimensional signals, like images and our tactile array, where the other bases were initially designed for one-dimensional signals and then expanded for two-dimensional signals.  A benefit of the contourlet transform  is that it is an \emph{overcomplete} dictionary, meaning its matrix representation $\basis_C \in \mR^{n\times p}$ has  $p > n$.  This generally allows each $\tsig$ to have multiple representations $\sparse$, with some possibly sparser than others. As stated in Section~\ref{cs.sec}, sparser signals allow for greater compression, which improves data transmission rates.

We apply the candidate basis to each time step of the true signal in the data set describe in Section~\ref{cs_exp.sec}.
To compute the sparse signals for the DCT and wavelets, we apply the inverse transform, $\sparse = \basis^{-1}\tsig$.  We used the Matlab \texttt{dct} function for the DCT transform and the \texttt{daubcqf} function from the Rice Wavelet Toolbox~\cite{wavelets} to generate the Haar (also know as D2) and D4 transforms.  
Since the contourlet transform does not generate a unique sparse representation, we solve the following optimization problem to determine $\sparse$,
\begin{equation} \label{bp.eq}
 \underset{\sparse \in \mR^{^p}}{\text{minimize}}~\| \sparse \|_1 ~~~\text{subject to}~\basis \sparse = \tsig.
\end{equation}
This approach has been shown to yield signals that exhibit a high degree of sparsity~\cite{Rubinstein2010b}.  
We used the Matlab contourlet toolbox~\cite{contourlets} function \texttt{pdfbdec}, with \texttt{`pkva'} filters and parameter \texttt{nlevs = [5]}, to generate the contourlet transform and the function \texttt{spg\_bn} from SPGL1 to solve (\ref{bp.eq})~\cite{spgl1:2007}.
In general, these signals $\sparse$ are only approximately sparse. Therefore, we compute the approximate sparsity $\tilde{k}$ of $\sparse$ using (\ref{sparsity.eq}) with the threshold $\tau = 0.001$ to evaluate the usefulness of each basis.  

Table~\ref{tab:BasisSparsity} shows the mean and max $\tilde{k}$ for each object for each representation basis, with the smallest (i.e., best) indicated in bold. For four of the five objects, the contourlet transform has the smallest $\tilde{k}$,
indicating it yields the most sparse representations.  The one exception is the granola box, where the Haar transform has the smallest $\tilde{k}$ and the contourlet transform has the largest. We believe this is due to the straight uniform edges in the granola box signals. The D4 transform had very similar values of $\tilde{k}$ to the Haar transform.  For most cases, the DCT yielded signals that were less sparse. Based on these results, we select the contourlet transform  and the Haar transform as bases for the rest of the paper.



\begin{table}[ht]
\caption{The mean and max approximate sparsity $\tilde{k}$ for different representation bases, using a threshold $\tau = 0.001$.  The smallest (best) value for each object is in bold.}
\begin{center}
  \begin{tabular}{| c c | c | c | c | c | c |}
    \hline
    \textbf{Basis}& & \parbox{.6cm}{\vspace{.1cm}\centering\textbf{Golf Ball} \vspace{.05cm}} & \parbox{1.0cm}{\vspace{.1cm} \centering \textbf{Granola Box} \vspace{.05cm} } & \textbf{Drill} & \textbf{Cup} & \textbf{Clamp} \\ \hline
    \multirow{2}{*}{Haar (D2)} & mean & 95.9 & \textbf{95.3} & 176.4 & 353.2 & 310.0 \\
              & max  & 146  & \textbf{209} & 265 & 462 & 425 \\ \hline
    \multirow{2}{*}{D4}        & mean & 93.1 & 206.5 & 178.1 & 351.3 & 238.5 \\
              & max  & 152  & 357 & 250 & 451 & 329 \\ \hline
    \multirow{2}{*}{DCT}       & mean & 148.4 & 360.2 & 176.1 & 404.0 & 222.9 \\
              & max  & 203 & 602 & 235 & 522 & 291 \\ \hline
    \multirow{2}{*}{Contourlet} & mean & \textbf{17.4} & 426.4 & \textbf{111.6} & \textbf{164.4} & \textbf{139.9} \\
              & max  &  \textbf{23} & 658 & \textbf{144} & \textbf{221} & \textbf{185} \\ 
    \hline
  \end{tabular}
\label{tab:BasisSparsity}
\end{center}
\end{table}



\subsubsection{Measurement matrix $\measMat$} \label{matrix.sec}

As stated in Section~\ref{cs.sec}, it is necessary that the measurement matrix $\measMat$ be chosen so that $\measMat \basis$ satisfies the necessary mathematical properties (e.g., RIP).  
Additionally, the measurement scheme should be implementable in hardware.  
Recall our measurements are of the form $\meas = \measMat \nsig$. This means each measurement $\meas_i$ is a weighted sum of the noisy taxel values, where the weights, implementable as hardware gains, are determined by the corresponding row of the measurement matrix.  A common choice for $\measMat$ that satisfies $k$-RIP with high probability is an i.i.d. random matrix with values drawn from a Gaussian distribution.  
If we were to use this measurement matrix,  the value of every taxel would appear in every measurement, since, with probability 1, every component of $\measMat$ is non-zero. Additionally, every measurement would use a different random gain for every taxel. 
To implement this measurement scheme in a tactile skin, either the taxel values would need to be individually sampled and then compressed in software or every taxel would need the hardware to generate $m$ gains, and the $m$ measurements would have to be collected in series. This is impractical in large-scale tactile arrays.

We consider two categories of measurement matrices, sparse and separable, that are more amenable to efficient hardware implementations, while still satisfying the necessary mathematical properties.

\begin{figure}
\includegraphics[width=\linewidth]{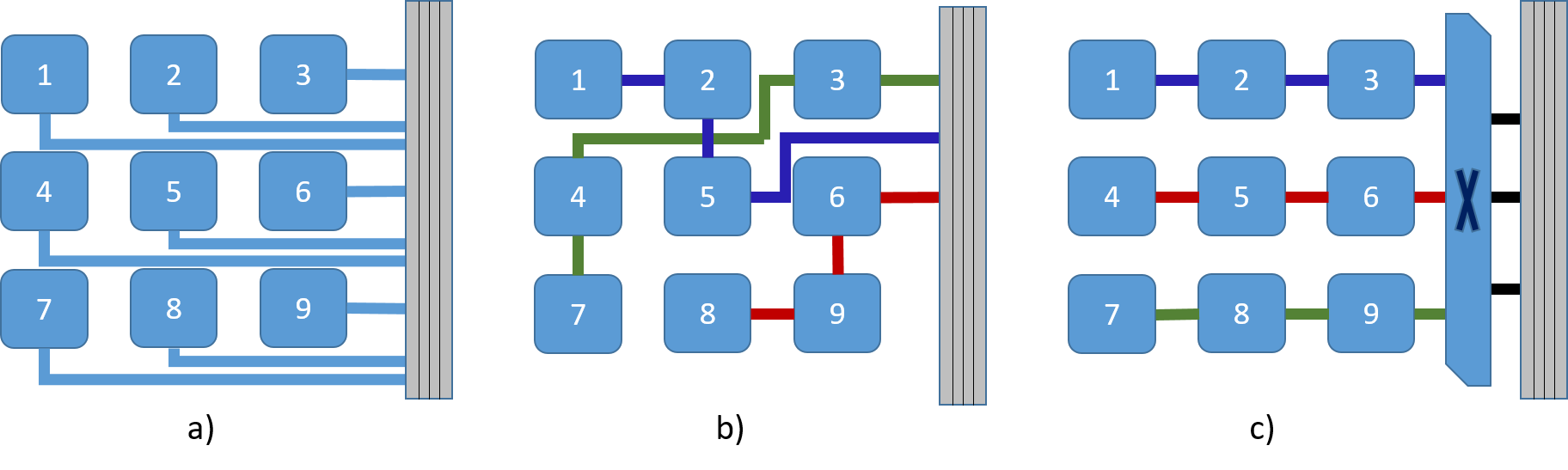}
\caption{Example wiring schematics for a) individual sensor measurements, b) sparse compressed sensing aggregated measurements, and c) separable compressed sensing aggregated measurements on a $3 \times 3$ tactile grid.}
\label{wiring.fig}
\end{figure}

\textbf{Sparse measurement matrices:} \label{sbhe.sec}
Sparse measurement matrices, i.e., matrices that contain a large number of zero-valued elements,  require that only a few taxel values be combined for each measurement. 
We call this set of taxels a \emph{measurement group}. 
To implement this scheme in hardware implementation, the taxels in each measurement group can be daisy-chained together.
Each measurement can be generated by performing a weighted sum across the taxels of the daisy-chain.  This could be accomplished by having each taxel's weighted value added to the weighted sum of the previous taxels in the chain and passing this new sum to the next taxel.  Or it could be done by aggregating all the taxels' values simultaneous, for example by having the taxels in the chain wired in parallel and measuring the resulting current.
Fig.~\ref{wiring.fig}b shows example daisy-chain wiring for sparse measurement matrix, which requires less wiring than wiring each taxel for individual sampling, as shown in Fig.~\ref{wiring.fig}a.

For our sparse measurement matrix, we select the Scrambled Block Hadamard Ensemble (SBHE) developed by Gan \emph{et al.}~\cite{gan08sbhe}.
They prove that for many basis matrices $\basis$ that have applications in image compression,
the product $\measMat \basis$ behaves like a Gaussian i.i.d. matrix, thus satisfying $k$-RIP.
SBHE is a partial block Hadamard transform with randomly permuted columns. It can be represented as 
\begin{equation}
\measMat_H = Q_mWP_n,
\end{equation}
where $W$ is a $n \times n$ block diagonal matrix with each block a $B \times B$ Hadamard matrix, $P_n$ is the permutation matrix that randomly reorders the $n$ columns of $W$, and $Q_m$ uniformly at random selects $m$ rows of $WP_n$.  Since the non-zero elements of $\measMat_H$ come from the $B \times B$ Hadamard blocks in $W$, $B$ determines the size of each measurement group. From a wiring standpoint, we want a small $B$. We find that $B = 32$ works well for our scenarios.  

SBHE has two additional benefits.  First, all non-zero elements are either $+1$ or $-1$.  This simplifies the hardware implementation since taxels only need the additional hardware to negate their values rather than hardware to generate different gains for each measurement.
Second, the measurement groups are disjoint subsets of taxels with each measurement group generating multiple measurements. This means that no wires are required between the measurement groups, allowing for disjoint daisy-chains.  This is beneficial as it means there is no need for coordination between measurement groups, which reduces system complexity and enables different measurements groups to take measurements in parallel.
Additionally it means each measurement group could use a single wire that is reused for each measurement, meaning less total wiring than other sparse systems.

\textbf{Separable measurement matrix:}
A separable measurement matrix can be written as $\measMat = (\measMat_1 \otimes \measMat_2)^T$, where $\otimes$ is the kronecker product.
In this case, (\ref{system.eq}) can be equivalently expressed as
\begin{equation} \label{separable.eq}
    Y(t) = \measMat_1^T X(t) \measMat_2,
\end{equation}
where $X(t)\in \mR^{\sqrt{n}\times\sqrt{n}}$  and $Y(t) \in \mR^{m_1 \times m_2}$, with $m = m_1m_2$, are matrix representations of the true signal $\tsig(t)$ and compressed signal $\meas(t)$ for time $t$.

This measurement scheme separates the compressed signal acquisition into two steps.  The first step, corresponding to $X(t) \measMat_2$,  generates linear combinations of the rows of $X(t)$; and thus for hardware, each row of taxels can be daisy-chained, as shown in Fig. \ref{wiring.fig}c. We refer to each row of taxels as a measurement group. The second step, the application of  $\measMat_1^T$, generates linear combinations of the columns of the matrix resulting from step 1. This requires additional hardware, which is represented in Fig.~\ref{wiring.fig}c by the component between the taxels and the network bus. To generate the measurements, a single column of $\measMat_2$ is applied across the measurement groups, computing a weighted sum of their values. Then $\measMat_1^T$ is applied to generate $m_1$ measurements. This is then repeated with a different column of $\measMat_2$ until all $m_2$ columns of $\measMat_2$ are used; thus generating the total $m=m_1\times m_2$ measurements per time step. Note, in general, every measurement incorporates every taxels' value.

We use a sub-sampled two-dimensional noiselet matrix $\measMat_{N2D} = (\measMat_{N1D} \otimes \measMat_{N1D})^T$, where $\measMat_{N1D} \in \mR^{\sqrt{n}\times\sqrt{n}}$ is the one-dimensional real-value noiselet matrix~\cite{Coifman2001}.  There are theoretical guarantees from compressed sensing that measurements with the noiselet matrix can recover signals sparse in the wavelet basis~\cite{Robucci2010,cs}. Further, it has successfully been used for image compressed sensing~\cite{Robucci2010}. Our noiselet matrix elements are either $+1$ or $-1$, similar to SBHE, thus for hardware implementation, noiselets also only need additional hardware to negate their values, compared to hardware to generate multiple various gains that would may be necessary for other separable matrices.

To achieve compression, we only transfer a subset of the measurements generated from applying $\measMat_1^T$.  This means, in general, our noiselet measurements do not reduce the sampling time, but the overall acquisition time is reduced due to transferring less data.  This is highly beneficial, as seen in the tactile system of Mukai \emph{et al.}~\cite{Mukai2008}. Their system locally samples at $1kHz$, but if they wish to acquire the full sensor signal at the processing center, their acquisition rate is only about $10Hz$. Therefore, by reducing the amount of data transmitted to the processing center, the tactile acquisition rate can be dramatically increased.

For our evaluations we used the \texttt{realnoiselet} function from Asif~\cite{noiselet} to create the one-dimensional noiselet matrix $\measMat_{N1D}$.


\subsubsection{Reconstruction algorithm} \label{FISTA.sec}

To solve problem (\ref{bpdn.eq}), we use an iterative, gradient-like method called Fast Iterative Shrinkage-Thresholding Algorithm (FISTA)~\cite{BT09}. 
We selected this algorithm because of its fast convergence; FISTA converges at a rate of $O(1/\kappa^2)$, where $\kappa$ is the iteration number, whereas standard gradient methods converge 
at a rate of $O(1/\kappa)$~\cite{BT09}.  In addition, we found it to be faster in practice in initial testing over other popular solvers, Alternating Directional Method of Multipliers~\cite{Boyd2011} and Smoothed L0~\cite{Mohimani2007}. 

Every time step we use FISTA to reconstruct the signal from the measurements taken in that time step.  As the true signal typically does not change much between time steps, we use the previous time step's reconstructed signal as the initial estimate for the current signal, which is called a \emph{warm-start}, and use FISTA to refine the estimate into a solution.
This warm-start led to faster convergence.
FISTA is also amenable to GPU implementations, which can further accelerate its performance.
We exploit this feature in our implementation.

We implemented FISTA in Matlab 2016a, utilizing Matlab's GPU capabilities to speed up the algorithm's execution as we assume an actual robot would have access to a GPU as part of its processing capabilities.   
We ran the algorithm for a constant number of iterations for each compressed sensing problem to standardize the reconstruction time. 

For each measurement-basis pair $A=\measMat\basis$, we set the FISTA stepsize parameter $L$ to twice the maximum eigenvalue of $A^TA$, which guarantees algorithm convergence~\cite{BT09}.
We then performed a grid search on the reconstruction of a few random time steps to set $\lambda$ (see (\ref{bpdn.eq})), which resulted in $\lambda = 0.1$ for SBHE measurements and $\lambda = 1$ for noiselet measurements.  We used these parameter values for the evaluations.
The reconstructed signal sometimes contained negative values, but since contact pressure must be non-negative, we set those elements to 0.

\subsection{Tactile Classification Solution} \label{class_solution.sec}

For tactile object classification, we apply compressed learning to utilize the compressed measurements from our data acquisition solution.  As these compressed measurements satisfy the compressed learning assumptions, the last necessary component of our solution is a multi-class classifier suitable for compressed learning.


Compressed learning was developed for the standard SVMs discussed in Section~\ref{svm.sec}, which are binary classifiers, but our problem requires multi-class classification.  Therefore instead of standard SVMs, we use the Direct Acyclic Graph SVM (DAGSVM)~\cite{Platt2000}, which is a multi-class SVM.  DAGSVM trains a binary SVM for each pair of classes.  During classification, an observation is classified by a binary SVM.  That observation is then classified in another binary SVM with the previously assigned class as one of the two potential classes and the previously rejected class eliminated from future consideration.  This is continued, sequentially eliminating classes from consideration, until a single class remains. The observation is classified as an element of the remaining class.  Thus for a $M$-class problem, DAGSVM trains $\frac{M(M-1)}{2}$ binary SVMs, but classifies an observation only using $M-1$ SVMs. It was found that the order in which the binary SVMs are used for classifying the observations does not matter~\cite{Platt2000}.  Since DAGSVM is essentially a combination of multiple binary SVMs, it is directly applicable to compressed learning.

In the following section, we show results from classification experiments to validate our approach.  In these experiments, we use the DAGSVM implemented in the MATLAB SVM Toolbox from University of East Anglia~\cite{Cawley2000} with a validation set to perform a grid search for the parameter $C$ in (\ref{softsvm.eq}).

\section{Evaluation} \label{eval.sec}
We evaluate our solutions using the data sets described in Section~\ref{data.sec}.

\subsection{Tactile Data Acquisition} \label{acq_results.sec}


For each time step $t$ of $\nsig(t)$ of the trajectories described in Section~\ref{cs_exp.sec}, we take compressed measurements $\meas(t) = \measMat \nsig(t)$.  We then use FISTA with a constant number of iterations and a warm-start to solve (\ref{bpdn.eq}) to obtain $\hat{\sparse}$.  Finally, we complete the recovery of the tactile signal by computing the reconstructed signal $\rsig = \basis \hat{\sparse}$.

We evaluate the quality of the reconstructed signals using the peak-signal-to-noise-ratio (PSNR) for each time step of the signal.  PSNR is a common metric for comparing a reconstructed signal or noisy signal to the original true signal, especially in images.  PSNR is calculated in decibels (dB) as:
\begin{equation}
    f_{PSNR}\Big(\rsig(t),\tsig(t),\mu\Big) = 10 \log_{10}{\frac{\mu^2}{\frac{1}{n}\|\tsig(t)-\rsig(t)\|_2^2}},
\end{equation}
where $\mu$ is the max value that an element in $\tsig$ can take, in our case 0.02.  Note, larger PSNRs correspond to more accurate reconstructions.

\subsubsection{Signal Compression} \label{compressioneval.sec}
We first evaluate our method by exploring the amount of signal compression we can achieve while obtaining high-quality reconstructed signals.  For SBHE measurements, the compression ratio does not affect the number of measurement groups, but it does reduce the number of measurements taken within each measurement.  This reduction can speed up the measurement process.  Additionally, the fewer measurements results in less data transferred across the network in each time step.  Finally, fewer measurements increase the reconstruction speed as the measurement matrix $\measMat$ is smaller, requiring fewer computations each iteration.  For noiselet measurements, the compression ratio has a minor affect on the number of measurements for each measurement group.  However, the improved compression ratio does reduce the amount of data that is  transferred on the network.  Also, similar to SBHE measurements, the reconstruction time is reduced due to the measurement matrix being smaller, which results in faster matrix multiplication.

We use three compression ratios, 3:1, 4:1, and 5:1, corresponding to 1365, 1024, and 819 measurements, respectively.  For each trajectory, we take measurements using both the SBHE and noiselet matrices.  For reconstruction, we use 20 FISTA iterations, and use both the Haar and contourlet transforms.  We evaluate the quality of the reconstructed signals using PSNR with respect to the true signal. This results in four sets of reconstructed signals for each trajectory.  
We also compute the PSNR for the sensor signal to provide a reference for the quality of the reconstructions.  

Table~\ref{tab:MResults} presents the mean and range of the PSNR of the four reconstructed signals and the sensor signal for each object's trajectory.  
As expected, the PSNR decreases, i.e., worsens, as the number of measurements decreases.   Except for the granola box, reconstructed signals from 1024 (4:1) measurements  with the contourlet transform have PSNR greater than or equal to the PSNR of the sensor signal; and, for the golf ball, drill, and cup, even using only 819 (5:1) measurements  with the contourlet transform has reconstructed signals with PSNR greater than or equal to the sensor signal.  For the Haar transform reconstructed signals, the golf ball, granola box, and drill have PSNR greater than or equal to the sensor signal for 1365 (3:1) measurements.
These results demonstrate that the reconstructed signals generated by our technique are better approximations of the true signals than would be obtained by sampling the sensor signal. 
For all the objects except the granola box, the contourlet transform reconstructions have higher PSNR than the Haar transform reconstructions.  For the granola box, the Haar transform achieves better PSNR.  The relative results between the Haar and contourlet transforms are reasonable given the approximate sparsities observed in Section~\ref{basis.sec}; the basis that provided sparser sparse signals for an object achieved the better reconstructed signals. 

\begin{table*}[ht]
\caption{Quality of Signal Reconstruction with Various Numbers of Measurements using 20 Iterations. }
\centering
\def\arraystretch{1.2} 
  \begin{tabular}{|c|c|c|c|c|c|c|c|c|} 
    \hline
    \multirow{3}{*}{\textbf{Object}} & \multirow{3}{*}{\parbox{1.5cm}{\centering \textbf{Measurement Matrix}}} & \multirow{3}{*}{\parbox{1.3cm}{ \centering \textbf{No. of Measurements}}} & \multicolumn{4}{|c|}{\textbf{Reconstructed Signal}} & \multicolumn{2}{|c|}{\textbf{Sensor Signal}} \\
    \cline{4-9}
     & & & \multicolumn{2}{|c|}{\textbf{Haar}} & \multicolumn{2}{|c|}{\textbf{Contourlet}}&  & \\
     \cline{4-7}
    & & & \parbox{1.5cm}{\vspace{.1cm} \centering \textbf{Mean PSNR} } & \parbox{1.5cm}{\vspace{.1cm} \centering \textbf{PSNR Range}} &  \parbox{1.5cm}{\vspace{.1cm} \centering \textbf{Mean PSNR}} & \parbox{1.5cm}{\vspace{.1cm} \centering \textbf{PSNR Range}}  &  \parbox{1.5cm}{\vspace{.1cm} \centering \textbf{Mean PSNR}} & \parbox{1.5cm}{\vspace{.1cm} \centering \textbf{PSNR Range}} \\ 
   \hline
    \multirow{6}{*}{
    	\begin{tabular}{@{}c@{}} Golf Ball \end{tabular}
    }
        & SBHE        & 
        \multirow{2}{*}{
    	\begin{tabular}{@{}c@{}} 1365 \end{tabular}
    	} & 31.4 & 29.4 - 34.0 & 36.6 & 35.1 - 38.9 & \multirow{6}{*}{
                                                	  \begin{tabular}{@{}c@{}} 29.0 \end{tabular} }
                                                	&
                                                	  \multirow{6}{*}{
                                                	  \begin{tabular}{@{}c@{}} 28.5 - 29.6 \end{tabular}    } \\
        & Noiselet    &  & 31.2 & 30.0 - 33.3 & 35.9 & 34.8 - 37.5 & & \\
        & SBHE        & 
        \multirow{2}{*}{
    	\begin{tabular}{@{}c@{}} 1024 \end{tabular}
    	} & 31.1 & 29.5 - 34.4 & 36.6 & 35.1 - 39.1 & & \\ 
	    & Noiselet    &      & 31.1 & 29.3 - 33.9 & 36.4 & 35.0 - 38.8 & &          \\
        & SBHE        &  
        \multirow{2}{*}{
    	\begin{tabular}{@{}c@{}} 819 \end{tabular}
    	} & 31.1 & 29.3 - 35.4 & 36.8 & 34.8 - 41.6 & & \\ 
        & Noiselet    &  & 30.9 & 28.8 - 34.2 & 36.5 & 34.7 - 39.5 & &                  \\ \hline
	\multirow{6}{*}{
        \begin{tabular}{@{}c@{}} Granola Box \end{tabular}
    }
    	& SBHE        & 
                \multirow{2}{*}{
            	\begin{tabular}{@{}c@{}} 1365 \end{tabular}
            	} & 30.2 & 28.0 - 34.1 & 26.1 & 23.9 - 39.0 & \multirow{6}{*}{
                                                        	  \begin{tabular}{@{}c@{}} 28.2 \end{tabular} }
                                                        	&
                                                        	  \multirow{6}{*}{
                                                        	  \begin{tabular}{@{}c@{}} 27.7 - 29.4 \end{tabular}    } \\
        & Noiselet    &  & 30.0 & 27.9 - 33.2 & 26.7 & 24.5 - 37.5 & & \\
        & SBHE        & 
        \multirow{2}{*}{
    	\begin{tabular}{@{}c@{}} 1024 \end{tabular}
    	} & 29.5 & 26.7 - 34.4 & 21.4 & 17.5 - 39.1 & & \\ 
        & Noiselets   &   & 29.4 & 26.4 - 34.0 & 20.6 & 16.9 - 38.7 & &                  \\
        & SBHE        &  
        \multirow{2}{*}{
    	\begin{tabular}{@{}c@{}} 819 \end{tabular}
    	} & 28.7 & 25.5 - 35.0 & 17.8 & 14.5 - 41.2 & & \\ 
        & Noiselets   &    & 28.7 & 25.5 - 34.6 & 18.2 & 14.9 - 39.6 & &                  \\ \hline
	\multirow{6}{*}{
	    \begin{tabular}{@{}c@{}} Drill \end{tabular}
	}
	    & SBHE        & 
        \multirow{2}{*}{
    	\begin{tabular}{@{}c@{}} 1365 \end{tabular}
    	} & 30.0 & 28.1 - 33.9 & 32.9 & 31.1 - 38.7 & \multirow{6}{*}{
                                                	  \begin{tabular}{@{}c@{}} 28.7 \end{tabular} }
                                                	&
                                                	  \multirow{6}{*}{
                                                	  \begin{tabular}{@{}c@{}} 28.1 - 29.5 \end{tabular}    } \\
        & Noiselet    &      & 29.8 & 28.0 - 33.5 & 33.1 & 31.5 - 37.5 & & \\
        & SBHE        & 
        \multirow{2}{*}{
    	\begin{tabular}{@{}c@{}} 1024 \end{tabular}
    	} & 29.3 & 27.1 - 34.3 & 31.9 & 29.7 - 39.3 & & \\ 
        & Noiselet       &  & 29.4 & 27.4 - 34.0 & 32.1 & 29.5 - 38.7 & &                  \\
        & SBHE        & 
        \multirow{2}{*}{
    	\begin{tabular}{@{}c@{}} 819 \end{tabular}
    	} & 28.8 & 26.4 - 35.5 & 30.3 & 27.1 - 41.3 & & \\ 
        & Noiselet       &   & 28.9 & 26.8 - 34.3 & 30.9 & 28.2 - 39.7 & &                  \\ \hline
	\multirow{6}{*}{
	    \begin{tabular}{@{}c@{}} Cup \end{tabular}
	}
        & SBHE        & 
        \multirow{2}{*}{
    	\begin{tabular}{@{}c@{}} 1365 \end{tabular}
    	} & 28.0 & 26.0 - 34.2 & 32.6 & 30.7 - 38.7 & \multirow{6}{*}{
                                                	  \begin{tabular}{@{}c@{}} 28.8 \end{tabular} }
                                                	&
                                                	  \multirow{6}{*}{
                                                	  \begin{tabular}{@{}c@{}} 28.2 - 29.5 \end{tabular}    } \\
        & Noiselet    &      & 28.1 & 26.3 - 33.5 & 32.6 & 31.1 - 37.6 & & \\
        & SBHE        & 
        \multirow{2}{*}{
    	\begin{tabular}{@{}c@{}} 1024 \end{tabular}
    	} & 27.1 & 24.9 - 34.1 & 31.7 & 29.8 - 39.2 & & \\ 
    	& Noiselets        & & 27.2 & 25.0 - 33.9 & 31.6 & 29.3 - 38.8 & &                  \\
    	& SBHE        & 
        \multirow{2}{*}{
    	\begin{tabular}{@{}c@{}} 819 \end{tabular}
    	} & 26.4 & 24.0 - 35.2 & 30.3 & 27.9 - 41.6 & & \\
	            & Noiselets        &  & 26.3 & 23.8 - 34.4 & 30.8 & 28.3 - 39.6 & & \\ \hline
	\multirow{6}{*}{
	    \begin{tabular}{@{}c@{}} Clamp \end{tabular}
	}
    	& SBHE        & 
        \multirow{2}{*}{
    	\begin{tabular}{@{}c@{}} 1365 \end{tabular}
    	} & 28.7 & 26.9 - 34.0 & 31.9 & 29.9 - 38.7 & \multirow{6}{*}{
                                                	  \begin{tabular}{@{}c@{}} 28.6 \end{tabular} }
                                                	&
                                                	  \multirow{6}{*}{
                                                	  \begin{tabular}{@{}c@{}} 28.0 - 29.5 \end{tabular}    } \\
        & Noiselet    &      & 28.6 & 27.0 - 33.6 & 32.1 & 30.4 - 37.4 & & \\
        & SBHE        & 
        \multirow{2}{*}{
    	\begin{tabular}{@{}c@{}} 1024 \end{tabular}
    	} & 27.9 & 25.8 - 34.5 & 30.4 & 27.7 - 39.3 & & \\ 
    	& Noiselets        &  & 27.8 & 25.9 - 33.8 & 30.3 & 27.8 - 38.8 & &                  \\
	    & SBHE        & 
        \multirow{2}{*}{
    	\begin{tabular}{@{}c@{}} 819 \end{tabular}
    	} & 27.4 & 25.1 - 35.1 & 28.4 & 25.3 - 41.2 & & \\
	    & Noiselets        &   & 27.3 & 24.8 - 34.4 & 29.0 & 25.8 - 39.8& &                  \\ \hline
  \end{tabular}
\label{tab:MResults}
\end{table*}

\begin{figure}[ht]
    \centering
    \includegraphics[width=\linewidth]{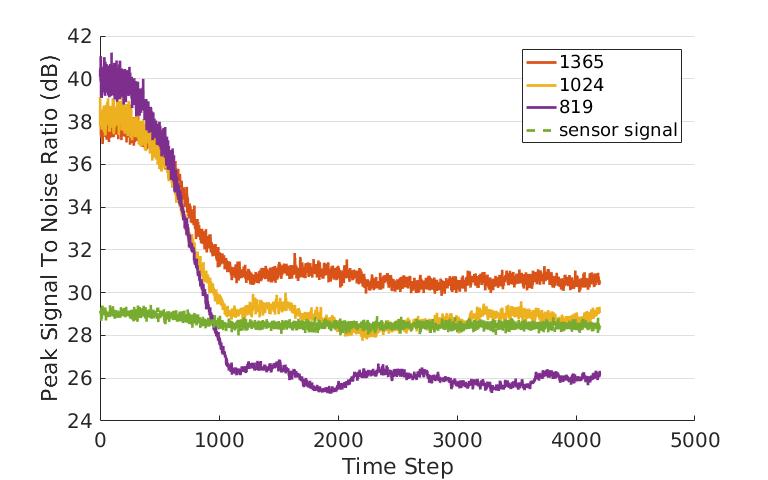}
    \caption{Peak Signal to Noise Ratio (PSNR) for different numbers of SBHE measurements (1365, 1024, and 819) corresponding to different compression ratios (3:1, 4:1, and 5:1) for the clamp.}
    \label{meas.fig}
\end{figure}

Fig.~\ref{meas.fig} shows the PSNR of each time step of the reconstructed signals, using the contourlet transform, for the clamp with 1365, 1024, and 819 (3:1, 4:1, and 5:1) measurements taken using the SBHE measurement matrix.  For all compression ratios, for approximately the first 1000 time steps, the PSNR decreases and then remains relatively stable. This pattern was observed for all reconstructed signals of all objects.  Those first 1000 time steps are when the tactile array comes down upon the object and the rest corresponds to the tactile array moving across the object.  The plot also shows that initially, the reconstructed signals from fewer measurements is better; though, the difference between the respective PSNR values is smaller than during the rest of the trajectory.  Fewer measurements likely have better PSNR at the beginning of the trajectory due to the relatively small signal compared to the noise.  The reasoning behind this assumption is the noise overwhelms the signal which leads to the reconstructed signals essentially reconstructing noise.  So, more measurements allow the noise to be reconstructed better, which makes the signal less like the true signal. 

\begin{figure*}[ht]
    \centering
    \includegraphics[width=.95\linewidth]{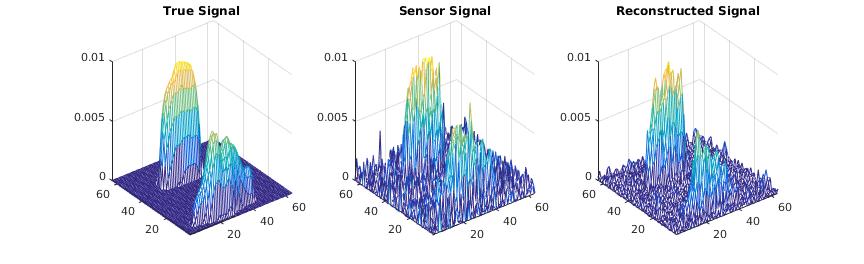}
    \caption{Taxel values from a $64 \times 64$ tactile array (4096 taxels) for the true signal $\tsig$ (left), sensor signal $\nsig$ (middle) and the reconstructed signal $\rsig$ (right). The reconstruction was from 1024 (4:1) SBHE measurements using the contourlet transform.  The reconstructed signal has a PSNR of 30.5 and the sensor signal has a PSNR of 28.7.}
    \label{array.fig}
\end{figure*}

We present a sample reconstructed signal tactile image, along with the true signal and sensor signal images in Fig.~\ref{array.fig}.  The signal comes from the drill trajectory with 1024 (4:1) measurements taken using the SBHE matrix and reconstructed with the contourlet transform.  While noise is still present, the reconstructed signal has a similar shape and magnitude to the true signal.  Furthermore the magnitude of the reconstructed signal's noise appears smaller than that of the sensor signal.

\subsubsection{Reconstruction Speed}
Our second experiment evaluates the reconstruction speed of our solution.  Faster reconstruction enables the robot to respond more quickly to the events causing the tactile signals.  We specifically look at the affect of the number of FISTA iterations.  The number of iterations is directly related to the speed of the reconstruction, with fewer iterations corresponding to faster reconstruction, but it also affects the signal quality.

The results in Section~\ref{compressioneval.sec} show that the reconstruction quality is largely independent of the choice of measurement matrix (SBHE or noiselet), and we observed similar behavior in all experiments.  Thus, for this experiment, we only present results of reconstructions from the SBHE measurements.  Since our reconstructed signals from 1024 (4:1) measurements had approximately as good or better PSNR than the sensor signal, we use this compression ratio.  
We use three different numbers of FISTA iterations, 20, 10, and 5, to reconstruct the signal from 1024 (4:1) measurements.  FISTA is initialized using a warm-start.  The reconstruction is performed using both the Haar and contourlet transforms.  We evaluate the performance using PSNR of the reconstructed signals and the mean time for performing the reconstructions.  

Table~\ref{tab:IterResults} shows the mean time, mean PSNR, and PSNR range for the reconstructed signals for each object's trajectory.  It also shows the mean PSNR and PSNR range for the sensor signals.  As expected, the mean reconstruction time is shorter when using fewer iterations.  The mean time for a given number of iterations and a given representation basis is consistent across all object trajectories.  The mean time is less for reconstructions performed using the Haar transform than for the contourlet transform, which is to be expected as the contourlet transform is a larger matrix.  The time per iteration does decrease slightly with more iterations, but it general, it is approximately 0.33 milliseconds when using the Haar transform and 0.4 milliseconds when using the contourlet transform.

\begin{table*}[ht]
\caption{Quality and Time of Signal Reconstruction with Various Numbers of Iterations using 1024 (4:1) Measurements}
\hfill{}
  \begin{tabular}{| c | c | c | c | c | c | c | c | c | c |}
  \hline
    \multirow{3}{*}{\textbf{Object}}  & \multirow{3}{*}{\parbox{1.3cm}{ \centering \textbf{No. of Iterations}}} & \multicolumn{6}{|c|}{\textbf{Reconstructed Signal}} & \multicolumn{2}{|c|}{\textbf{Sensor Signal}} \\
    \cline{3-10}
    & & \multicolumn{3}{|c|}{\textbf{Haar}} & \multicolumn{3}{|c|}{\textbf{Contourlet}}&  & \\
     \cline{3-8}
    & & \parbox{1.2cm}{\vspace{.1cm} \centering \textbf{Mean Time} } & \parbox{1.2cm}{\vspace{.1cm} \centering \textbf{Mean PSNR} } & \parbox{1.2cm}{\vspace{.1cm} \centering \textbf{PSNR Range}} & \parbox{1.2cm}{\vspace{.1cm} \centering \textbf{Mean Time} } &  \parbox{1.2cm}{\vspace{.1cm} \centering \textbf{Mean PSNR}} & \parbox{1.2cm}{\vspace{.1cm} \centering \textbf{PSNR Range}}  &  \parbox{1.2cm}{\vspace{.1cm} \centering \textbf{Mean PSNR}} & \parbox{1.2cm}{\vspace{.1cm} \centering \textbf{PSNR Range}} \\ 
   \hline
    \multirow{3}{*}{
    	\begin{tabular}{@{}c@{}} Golf Ball \end{tabular}
    }
    & 20 & 6.4 & 31.1 & 29.3 - 33.9 & 7.7 & 36.4 & 35.0 - 38.8  & \multirow{3}{*}{
                                                            	  \begin{tabular}{@{}c@{}} 29.0 \end{tabular} }
                                                            	&
                                                            	  \multirow{3}{*}{
                                                            	  \begin{tabular}{@{}c@{}} 28.5 - 29.6 \end{tabular}    } \\
    & 10 & 3.3 & 31.4 & 29.9 - 34.7 & 4.0 & 37.9 & 36.6 - 41.0 & &                  \\
    & 5 & 1.8 & 31.9 & 30.6 - 37.5 & 2.1 & 39.5 & 38.2 - 46.3 & &                 \\ \hline
    \multirow{3}{*}{
    	\begin{tabular}{@{}c@{}} Granola Box \end{tabular}
    }
    & 20 & 6.4 & 29.4 & 26.4 - 34.0 & 7.7 & 20.6 & 16.9 - 38.7  & \multirow{3}{*}{
                                                        	  \begin{tabular}{@{}c@{}} 28.2 \end{tabular} }
                                                        	&
                                                        	  \multirow{3}{*}{
                                                        	  \begin{tabular}{@{}c@{}} 27.7 - 29.4 \end{tabular}    } \\
    & 10 & 3.3 & 30.0 & 27.2 - 34.3 & 4.0 & 20.9 & 17.4 - 40.6 & & \\
    &  5 & 1.8 & 30.6 & 27.4 - 37.3 & 2.1 & 21.1 & 17.1 - 45.4  & &                  \\ \hline
    \multirow{3}{*}{
    	\begin{tabular}{@{}c@{}} Drill \end{tabular}
    }
    & 20 & 6.4 & 29.4 & 27.4 - 34.0 & 7.7 & 32.1 & 29.5 - 38.7 & \multirow{3}{*}{
                                                	  \begin{tabular}{@{}c@{}} 28.7 \end{tabular} }
                                                	&
                                                	  \multirow{3}{*}{
                                                	  \begin{tabular}{@{}c@{}} 28.1 - 29.5 \end{tabular}    } \\
    & 10 & 3.3 & 29.8 & 28.0 - 34.7 & 4.0 & 33.2 & 30.6 - 40.6& &                  \\
    &  5 & 1.8 & 30.3 & 28.5 - 37.6 & 2.1 & 34.2 & 31.4 - 45.6& &  \\ \hline
    \multirow{3}{*}{
    	\begin{tabular}{@{}c@{}} Cup \end{tabular}
    }
    & 20 & 6.4 & 27.2 & 25.0 - 33.9 & 7.6 & 31.6 & 29.3 - 38.8 & \multirow{3}{*}{
                                                	  \begin{tabular}{@{}c@{}} 28.8 \end{tabular} }
                                                	&
                                                	  \multirow{3}{*}{
                                                	  \begin{tabular}{@{}c@{}} 28.2 - 29.5 \end{tabular}    } \\
    & 10 & 3.3 & 27.6 & 25.5 - 35.3 & 4.1 & 32.4 & 30.1 - 40.5& &                  \\
    &  5 & 1.8 & 27.9 & 25.9 - 38.2 & 2.1 & 33.0 & 30.6 - 45.5 & &   \\ \hline
    \multirow{3}{*}{
    	\begin{tabular}{@{}c@{}} Clamp \end{tabular}
    }
    & 20 & 6.3 & 27.8 & 25.9 - 33.8 & 7.6 & 30.3 & 27.8 - 38.8 & \multirow{3}{*}{
                                                	  \begin{tabular}{@{}c@{}} 28.6 \end{tabular} }
                                                	&
                                                	  \multirow{3}{*}{
                                                	  \begin{tabular}{@{}c@{}} 28.0 - 29.5 \end{tabular}    } \\
    & 10 & 3.3 & 28.2 & 26.4 - 34.7 & 4.0 & 31.3 & 28.8 - 40.7 & &                  \\
    &  5 & 1.8 & 28.7 & 26.9 - 37.5 & 2.2 & 32.2 & 29.6 - 45.4 & &  \\       
    		    \hline
  \end{tabular}
\hfill{}
\label{tab:IterResults}
\end{table*}

The PSNR between reconstructed signals using the Haar and contourlet transforms follow the same trend as in Table~\ref{tab:MResults}. Contrary to expectations, we see in Table~\ref{tab:IterResults} that the PSNR of the reconstructed signals increases as the number of iterations, and thus time, decreases.  This is likely because of the warm-start, which initializes FISTA with a good approximation of the solution, and because FISTA is solving (\ref{bpdn.eq}), which optimizes with respect to the sensor signal (and signal sparsity), not the true signal, but we evaluate the reconstruction with respect to the true signal since it is the information we actually want.  In other words, if we could optimize with respect to the true signal, the PSNR would decrease (or stay the same) as the number of iterations decrease.

\begin{figure}[ht]
    \centering
    \includegraphics[width=\linewidth]{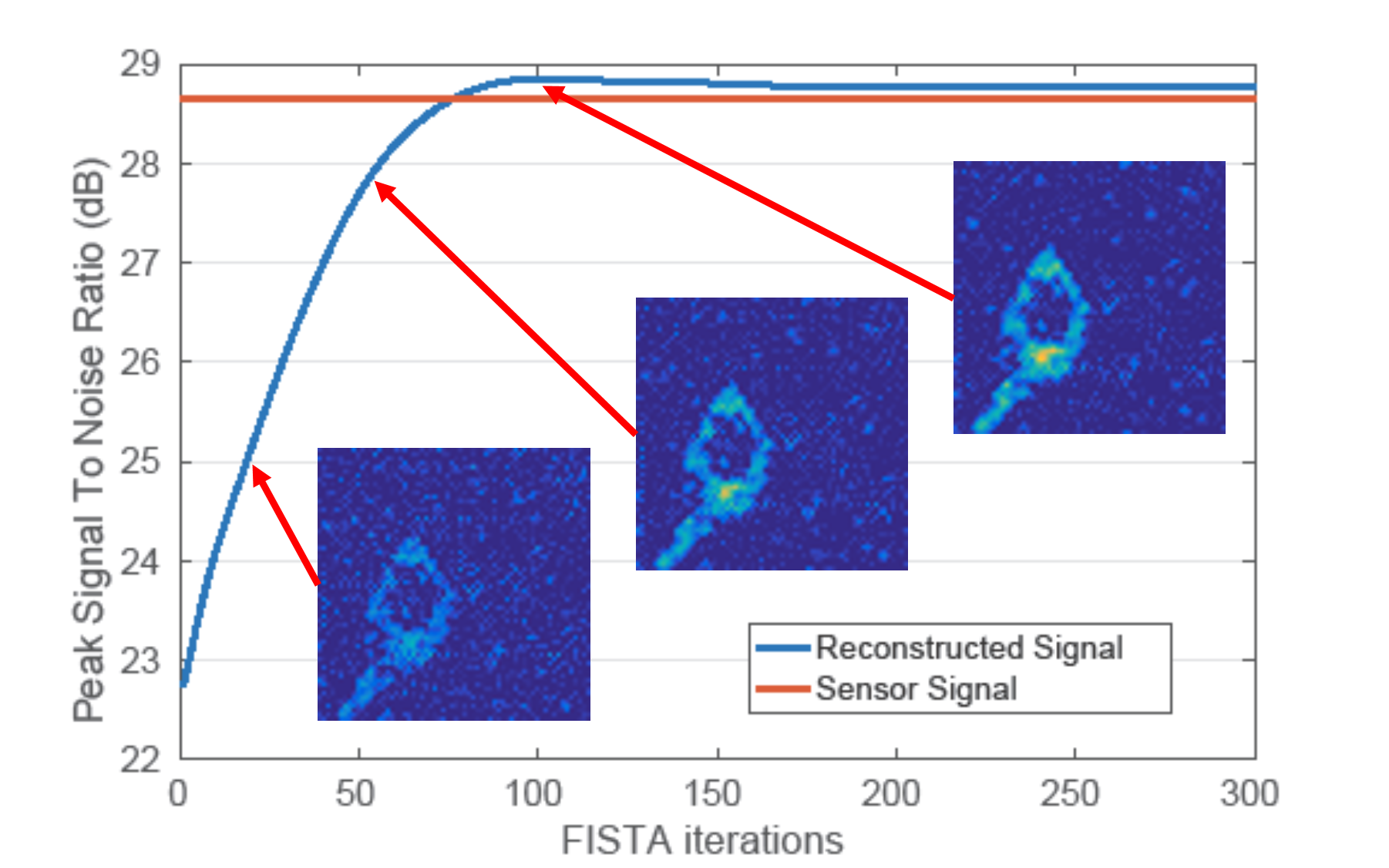}
    \caption{PSNR for different numbers of FISTA iterations.  The visualized reconstructed signals correspond to 20, 50, and 100 FISTA iterations.} 
    \label{converge.fig}
\end{figure}

Part of the reason we are able to use so few FISTA iterations is our use of a warm-start and the fact that the signals do not change much between time steps.  We also looked at the convergence of the PSNR without a warm-start.  Fig.~\ref{converge.fig} shows the PSNR for a reconstructed signal of the clamp against the numbers of FISTA iterations when reconstructing the signal without a warm-start (i.e., initializing all the signal elements to zero). The reconstruction is from 1024 (4:1) SBHE measurements using the contourlet transform.  It also shows a visualization of the reconstructed signal's tactile image at 20, 50, and 100 FISTA iterations. After about 100 iterations, the PSNR converges, but even with only 20 iterations, the tactile image shows the reconstructed signal contains the shape, though not magnitude, seen after 100 iterations. 
This shows that a warm-start does improve our reconstruct speed for slow changing signals, but even without a warm-start we can reconstruct decent approximations to the true signal.

\subsection{Object Classification Results} \label{class_results.sec}


To evaluate our compressed learning tactile object classification approach, we start with the data set of sensor signals from the array of $4096$ taxels and generate the compressed measurements.  
The development set, used for validation training as described in Section~\ref{svm.sec}, was formed by choosing a percentage of the 360 perturbations (i.e., starting position and orientation) described in Section~\ref{to_exp.sec} uniformly at random.  All observations with those perturbations for all 16 objects defined the development set.  The validation set, for validation testing to select the parameter $C$ of (\ref{softsvm.eq}), was formed similarly from the remaining perturbations.  The development set and validation set were then used with the DAGSVM described in Section~\ref{class_solution.sec} to train our classifier.  The remaining perturbations were used as our test set to evaluate the quality of the classifier by using the classifier on each of the observations and comparing the resulting label to the object used to generate the observation.
For comparison, we also performed classification using the data sets of sensor signals.  
For convention, we use \emph{signal size} to refer to the number of elements in a signal's vector.  For compressed signals, the signal size is $m$, the number of measurements, and for the sensor signals, the signal size is the number of taxels in the array.  


\subsubsection{Signal Size}
We first evaluate the accuracy of our classification with respect to the signal size.  Similar to tactile data acquisition, smaller signal size means fewer measurements and less data to transfer.  It also reduces the amount of data to process during classification, which reduces the computational needs of the training and classification processes.

We use data sets for classification with the following number of measurements $m$: 4096, 256, 64, 16, 4, and 1.  The compressed measurements were taken with both the SBHE and noiselet measurement matrices. We also use the sensor signals with the corresponding number of taxels.  We use $40\%$, i.e., 144, of the 360 perturbations for the development set, which results in the full development set containing 2304 observations.  The validation set is $20\%$ of the 360 perturbations, corresponding to 1152 observations, chosen randomly from observations not in the development set.  The test set is the remaining 2304 observations.

\begin{figure}
\includegraphics[width=\linewidth]{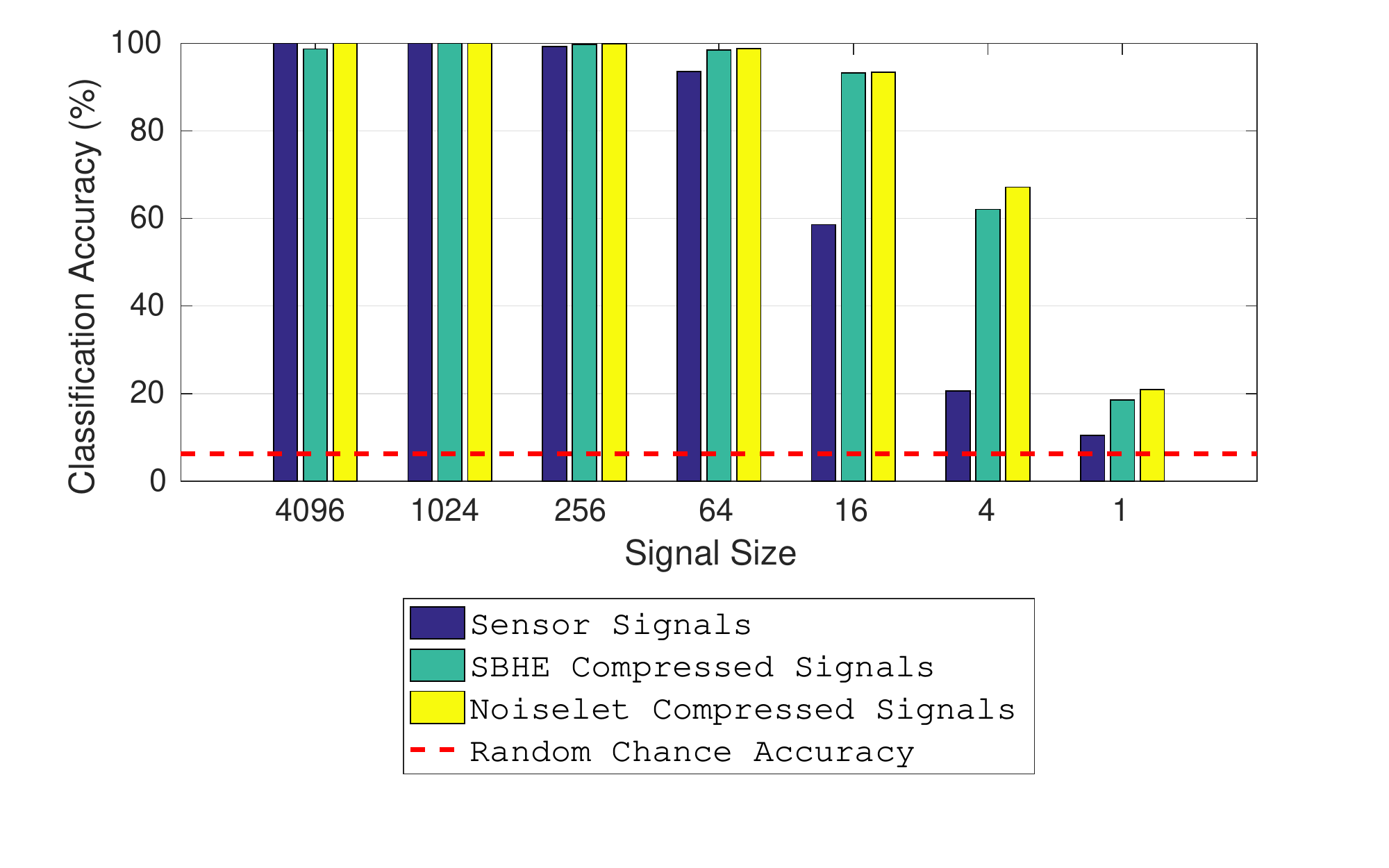}
\caption{The classification accuracy for various signal sizes using two training set sizes.  
For the sensor signals, signal size refers to the number of taxels in the array, and for the compressed signals, signal size refers to the number of measurements.  The dotted line is the accuracy of randomly assigning a label to each example.}
\label{signal_size.fig}
\end{figure}

Fig. \ref{signal_size.fig} shows the overall classification accuracies for seven different signal sizes.
In addition to the accuracy rates for the compressed signals, we also look at the accuracy rates for the sensor signals of corresponding dimensions. All the results are averages over 10 different splits of the data set into training (combined development and validation) and test sets. 
Smaller signal sizes yield less accurate classification, but even with fairly small signal sizes, the classification still has a high success rate. 
The sensor signals achieve over 93.3\% accuracy even at a signal size of 64.  The compressed signals achieve a similar level of accuracy for the signal size of 16, a compression ratio of 256:1, with 93.2\% and 93.4\% respectively for the SBHE and noiselet measurements.  Overall, the compressed signals outperformed the corresponding sensor signals of the same size.  The exception is for the SBHE compressed signals for which no actual compression was performed, i.e., SBHE `compressed' signals of signal size 4096.  In general, the different measurement matrices perform similarly.

The results agree with Theorem \ref{cl.thm}. The compressed signals have similar accuracies to the classifier on the original signal (the sensor signal of size 4096).  Further, the accuracy deviates more with increased compression.


\subsubsection{Training Set Size}
Collecting large sets of training data can be inconvenient, challenging, or impractical in deployed hardware.   
Therefore, we explore performance with respect to various amounts of training data to determine how much is needed for accurate classification.
We used development sets with $40\%, 20\%, 10\%, 6\%, 2\%$, and $0.66\%$ of the perturbations, with corresponding validation sets of $20\%, 5\%, 3\%, 2\%$, and $0.33\%$ of the perturbations.  We used a signal size of 16 measurements, using both SBHE and noiselet measurements.  We also train and classify with sensor signals of size 4096 and 16.

\begin{figure}
\includegraphics[width=\linewidth]{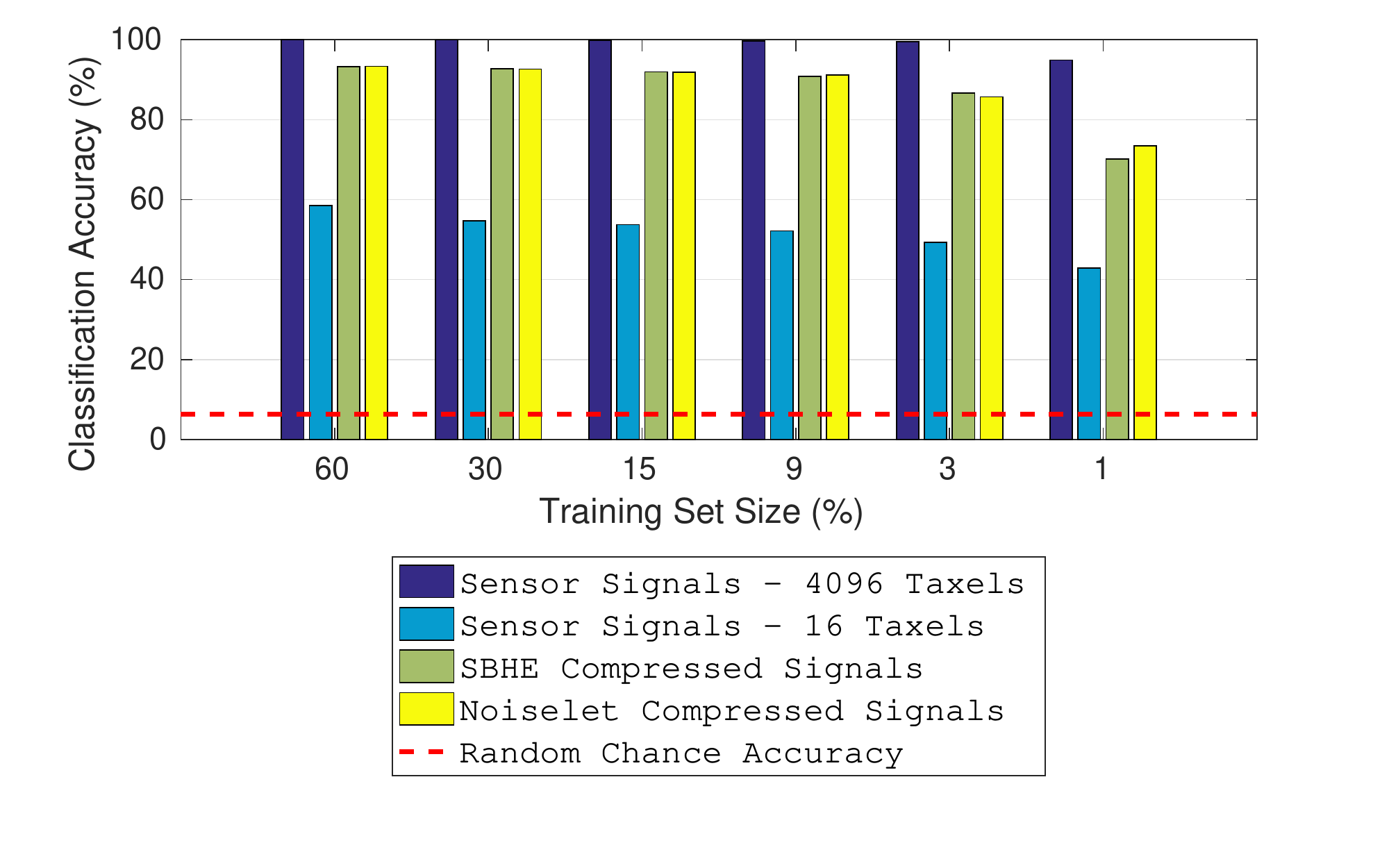}
\caption{The classification accuracy for training sets of various sizes.  The sensor signals of 4096 taxels are the signals used to generate the compressed signals.  The compressed signals have a signal size of 16, i.e., have 16 measurements per observation.
The dotted line is the accuracy of randomly assigning a label to each observation.}
\label{training_size.fig}
\end{figure}

Our results are shown in Fig. \ref{training_size.fig}.  It shows the classification rates for the sensor signals of size 4,096, compressed signals
obtained from these sensor signals, of size 16, and sensor signals obtained from coarser tactile arrays of size 16.  
As with Fig.~\ref{signal_size.fig}, the results are averaged over 10 splits of the data set.
Our method achieves high accuracy even with fairly small training sets, though the classification accuracy decreases as the amount of training data decreases.  The compressed signals of size 16 maintain over 80\% accuracy even with 3\% training data.  At 1\% training data, the compressed signals classification accuracy drops to approximately 70\%.  This is much better than sensor signals of size 16, which have under 60\% accuracy for all training sizes.

As was the case for the results shown in Fig. \ref{signal_size.fig}, these results also agree with theoretical results from compressed learning.  The classification accuracies for the compressed signals continue to be similar to the results from the original signals, but the deviation increases as the training set size decreases in accordance with (\ref{bound.eq}).  This is most clearly seen in Fig. \ref{training_size.fig} by comparing the deviations between original signals (sensor signals of size 4,096) and the compressed signals of size 16.  The deviation goes from approximately 7\% to 15\% between 60\% and 1\% training set sizes.

\subsubsection{Per Object Classification}
\begin{figure}
\begin{tabular}{c}
\subfloat[SBHE] {\includegraphics[width = .5\textwidth]{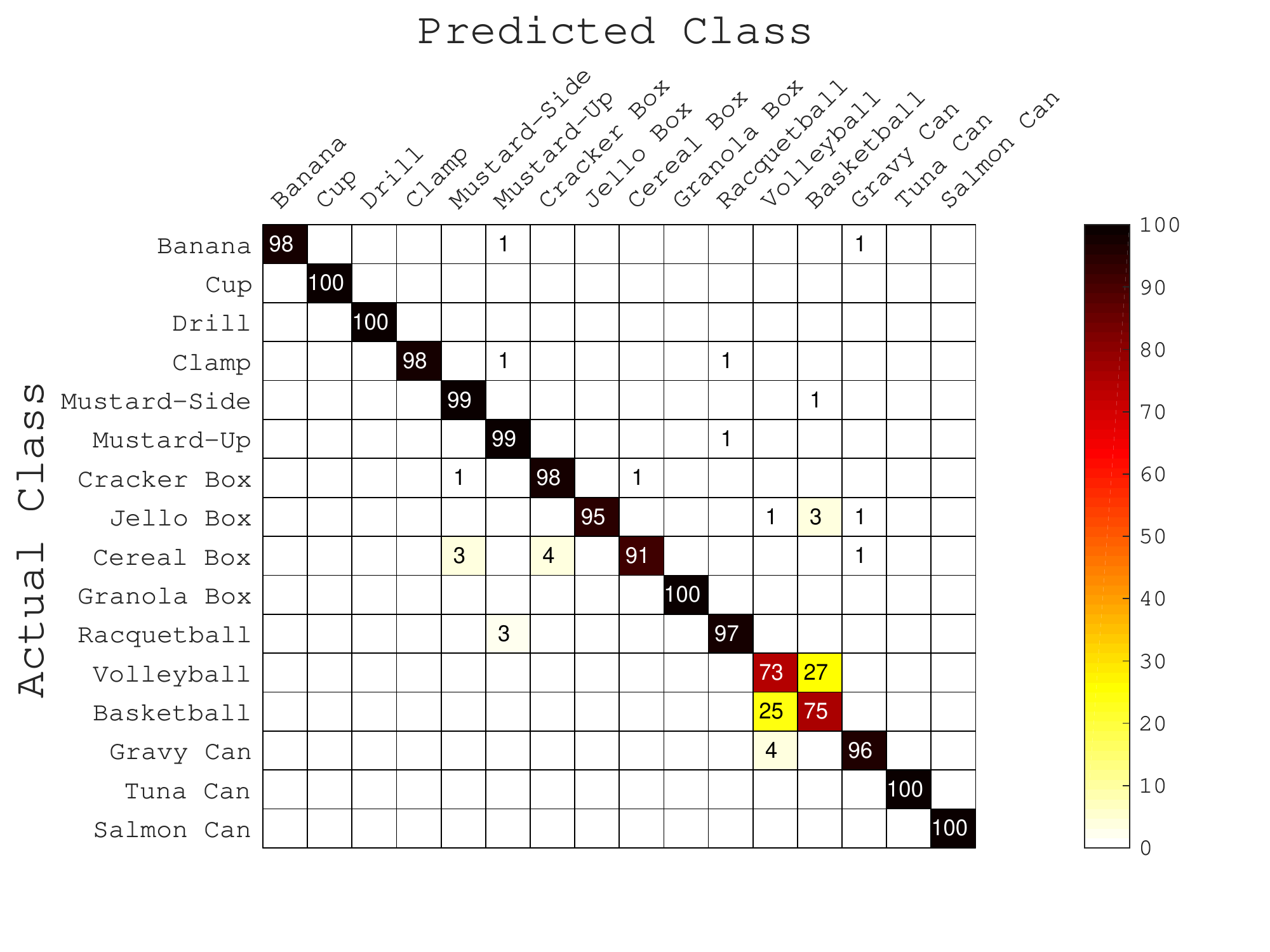}\label{cm_sbhe.fig}} \\
\subfloat[Noiselet]  {\includegraphics[width = .5\textwidth]{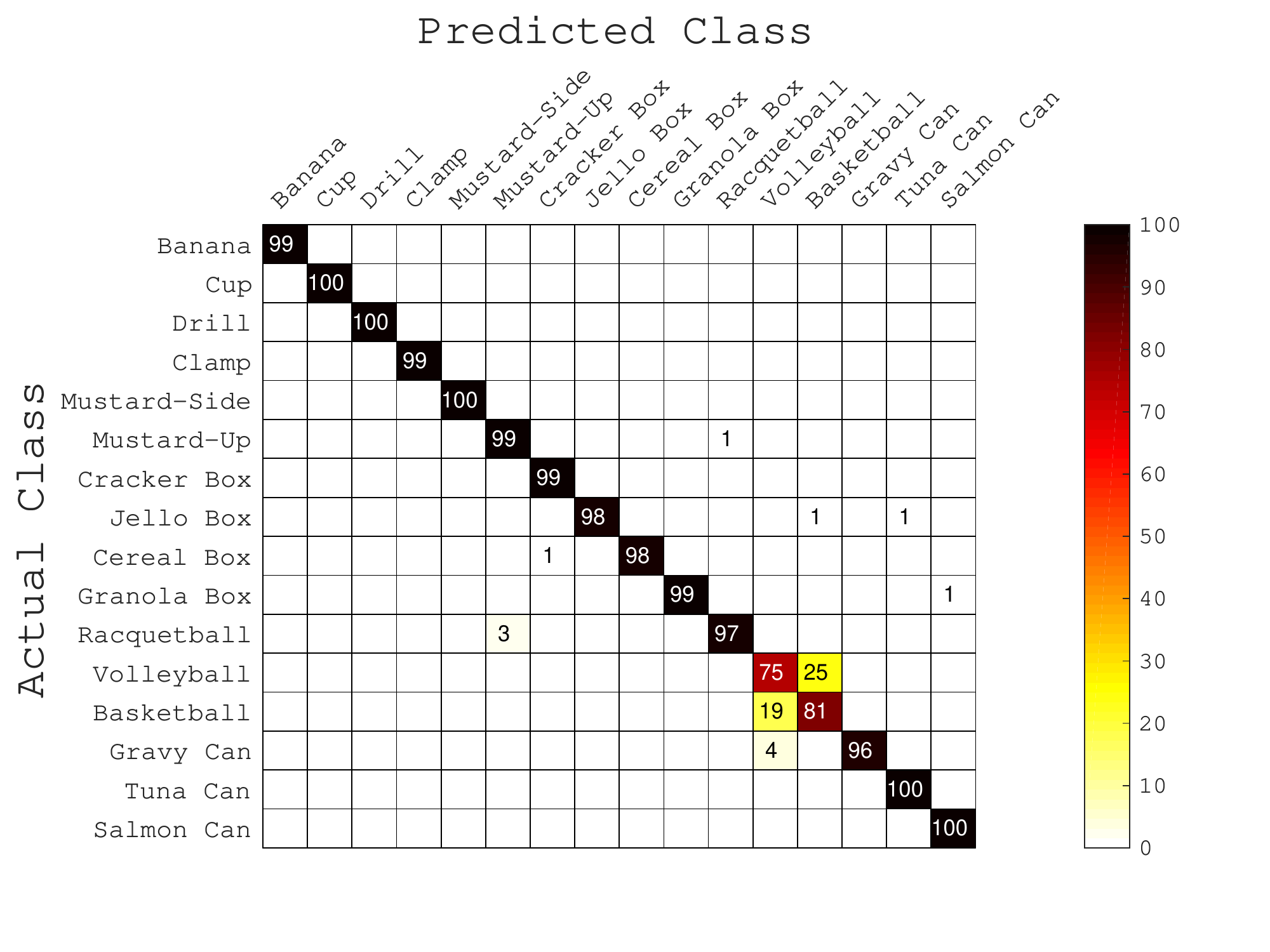}\label{cm_n.fig}}
\end{tabular}
\caption{The average confusion matrix over the 10 data-set splits for compressed signals of 64 elements trained on three percent of the examples of each object for (a) SBHE measurements and (b) noiselet measurements.  The values are the percentage of the actual class examples that were label as the predicted class.  Elements with no stated values have approximately zero percent of the actual class labeled as the predicted class.}
\label{confusion.fig}
\end{figure}

To get a better understanding of how the classifier performs among the classes, we computed the confusion matrix, which shows the percentage of observations of each class that are labeled as a particular class.  
Fig. \ref{confusion.fig} shows the confusion matrices for the (a) SBHE and (b) noiselet compressed signals of size 64, trained on 3\% of the observations per object, averaged over the ten splits of the data-set.  As both compressed signal types perform similarly, we discuss them together.  From the strong diagonal it is clear the classification performs well overall.
The greatest confusion occurs between the volleyball and the basketball; approximately 25\% of the time one is mistaken for the other.  This is understandable because both are spheres with similar radii and similar tactile signals, as seen in Figs. \ref{volleyball.fig} and \ref{basketball.fig}.  The gravy can and the volleyball also generate a bit of confusion.  While this is less intuitive, it is not surprising.  Both objects have round shapes, and while the volleyball has a much larger radius overall, Figs. \ref{volleyball.fig} and \ref{gravy.fig} show the contact radii are similar.  There is also a little confusion between the upright mustard bottle and the racquetball since they also have circular contacts of similar radii.  The other confusions of note are classifying the cereal box as either the cracker box or the mustard bottle on its side and the jello box as the basketball.  This is a little less apparent, but the shape and dimensions are similar between the cereal box and the other two items, and the basketball covers similar area as the jello box.

\section{Conclusion} \label{concl}
We have developed an approach for tactile data acquisition using compressed sensing.
In our approach, tactile data is compressed in hardware before transmission to the central processing unit.
Then, the full tactile data is reconstructed from the compressed data when needed.
Using standard compressed sensing tools, we achieved signal reconstruction of 4096 taxels at a rate on the order of $100Hz$, 
on par with the system presented by Schmitz \emph{et al.}~\cite{schmitz11icub}, which is currently the largest system by number of taxels.
Further, our system was able to compress the signal to a fourth of its original size and produce a quality reconstructed signal.  
Additionally, we have discussed how compressed sensing may provide guidance for new wiring techniques with potential to reduce wiring complexity. 
From this investigation,  we conclude that compressed sensing is a feasible approach for tactile data acquisition and worth continued development.
We believe this paradigm can open the door to larger-scale tactile systems as well as faster data acquisition.

In addition, we utilized the measured compressed signals directly to perform tactile object classification.  We were able to classify various objects with high accuracy, even with large amounts of compression and small amounts of training data.  Classification on compressed signals resulted in accuracy similar to the classification on the original sensor signals, and it outperformed classification using full sensor signals of comparable sizes to the compressed signals.  
Our approach offers benefits of reduced data acquisition and processing time, as well as the potential to reduce wiring complexity in hardware implementations.

Directions for future work include expanding compressed sensing ideas for other skin systems, for example, skins with different sensor layouts or multi-modal sensors.  
We also plan to integrate compressed sensing with other applications, for example, object manipulation.

\ifCLASSOPTIONcaptionsoff
  \newpage
\fi



\bibliographystyle{IEEEtran}
\end{document}